\documentclass[10pt,twocolumn,letterpaper]{article}

\usepackage[table]{xcolor}
\definecolor{lightgreen}{RGB}{240, 255, 240}  
\definecolor{headergray}{gray}{0.96}
\usepackage{pifont}  
\usepackage[pagenumbers]{cvpr} 

\definecolor{cvprblue}{rgb}{0.21,0.49,0.74}
\usepackage[pagebackref,breaklinks,colorlinks,allcolors=cvprblue]{hyperref}

\def\paperID{15248} 
\def\confName{CVPR}
\def\confYear{2026}
\usepackage{multirow}
\usepackage[table]{xcolor}
\definecolor{lightgray}{gray}{0.94}
\title{Fighting Hallucinations with Counterfactuals: Diffusion-Guided Perturbations for LVLM Hallucination Suppression}
\author{
Hamidreza Dastmalchi\\
\fontsize{10}{11} \selectfont York University\\
\fontsize{10}{11} \selectfont Toronto, Canada\\
{\tt\footnotesize hrd@yorku.ca}
\and
Aijun An\\
\fontsize{10}{11} \selectfont York University\\
\fontsize{10}{11} \selectfont Toronto, Canada\\
{\tt\footnotesize aan@yorku.ca}
\and
Ali Cheraghian\\
\fontsize{10}{11} \selectfont Macquarie University\\
\fontsize{10}{11} \selectfont Sydney, Australia\\
 {\tt\footnotesize ali.cheraghian@mq.edu.au}
\and
Hamed Barzamini\\
\fontsize{10}{11} \selectfont Northern Illinois University\\
\fontsize{10}{11} \selectfont Illinois, USA\\
{\tt\footnotesize h.barzamini@niu.edu }
}

\begin{document}
\maketitle
\begin{abstract}
While large vision–language models (LVLMs) achieve strong performance on multimodal tasks, they frequently generate hallucinations—unfaithful outputs misaligned with the visual input. To address this issue, we introduce CIPHER (Counterfactual Image Perturbations for Hallucination Extraction and Removal), a training-free method that suppresses vision-induced hallucinations via lightweight feature-level correction. Unlike prior training-free approaches that primarily focus on text-induced hallucinations, CIPHER explicitly targets hallucinations arising from the visual modality.
CIPHER operates in two phases. In the offline phase, we construct OHC-25K (Object-Hallucinated Counterfactuals, 25,000 samples), a counterfactual dataset consisting of diffusion-edited images that intentionally contradict the original ground-truth captions. We pair these edited images with the unchanged ground-truth captions and process them through an LVLM to extract hallucination-related representations. Contrasting these representations with those from authentic (image, caption) pairs reveals structured, systematic shifts spanning a low-rank subspace characterizing vision-induced hallucination.
In the inference phase, CIPHER suppresses hallucinations by projecting intermediate hidden states away from this subspace. Experiments across multiple benchmarks show that CIPHER significantly reduces hallucination rates while preserving task performance, demonstrating the effectiveness of counterfactual visual perturbations for improving LVLM faithfulness. Code and additional materials are available at \url{https://hamidreza-dastmalchi.github.io/cipher-cvpr2026/}.
\end{abstract}

\section{Introduction}
\label{sec:intro}
Large vision-language models (LVLMs) commonly integrate vision encoders with pretrained large language models (LLMs) via lightweight connector modules, enabling joint reasoning over visual and textual inputs. This modular design underpins models like MiniGPT-4~\cite{zhu2023minigpt}, LLaVA~\cite{LLaVA, LLaVA2}, and mPLUG-Owl2~\cite{ye2024mplug1, ye2024mplug2}, which are further refined through instruction tuning on multimodal datasets.

\begin{figure}[!t]
    \centering
    \includegraphics[width=\linewidth]{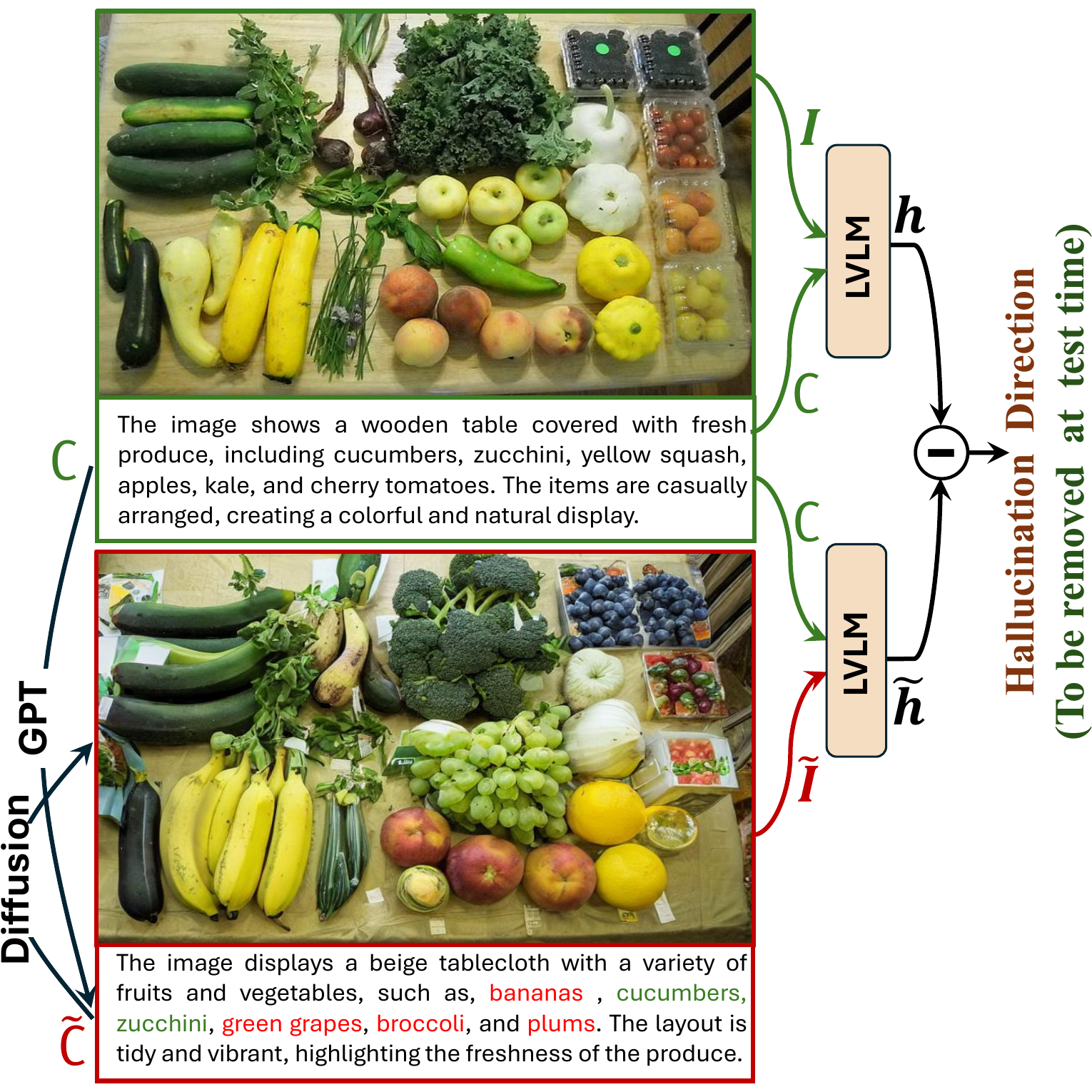}
\caption{Given an image--caption pair $(\boldsymbol{I}, \mathcal{C})$, we generate a counterfactual image $\tilde{\boldsymbol{I}}$ using a diffusion model conditioned on a GPT-perturbed caption $\tilde{\mathcal{C}}$ via controlled perturbations. A vision--language model encodes both $(\boldsymbol{I}, \mathcal{C})$ and $(\tilde{\boldsymbol{I}}, \mathcal{C})$, yielding features $\boldsymbol{h}$ and $\tilde{\boldsymbol{h}}$. Their difference $\boldsymbol{\delta} = \tilde{\boldsymbol{h}} - \boldsymbol{h}$ captures the hallucination direction we later nullify at inference.
}

    \label{fig:motivation}
    \vspace{-0.4cm}
\end{figure}

While LVLMs show remarkable capabilities, they often suffer from \textit{hallucination}, generating descriptions of objects, attributes, or scenes not present in the input image. Different studies have explored the roots of hallucination \cite{survey1, survey2}, showing that it arises not only from the generative tendencies of LLMs but also from weak vision grounding, modality misalignment, biased training data, and context-blind decoding strategies. 

Hallucination mitigation in LVLMs generally falls into three categories. (1) \textit{Training-based} methods improve visual grounding through additional supervision or objective design but require costly annotations, retraining, or architectural changes \cite{liu2023lrv, you2023ferret, liu2023mitigating, jiang2024hallucination, gunjal2024detecting, chen2024internvl}. (2) \textit{Post-processing} methods detect and revise hallucinations after generation using external tools \cite{zhou2023lure, yin2024woodpecker}. (3) \textit{Test-time} techniques are lightweight strategies applied during inference that adapt internal representations or the decoding process without requiring retraining or modifying model parameters \cite{huang2023opera, leng2024mitigating-vcd, liu2025reducing-vti, yang2025nullu}. Although effective, some test-time approaches, including contrastive decoding, require multiple forward passes \cite{leng2024mitigating-vcd}, leading to higher inference cost. In contrast, test-time feature-level intervention techniques are more efficient, requiring only a single forward pass \cite{yang2025nullu}. However, existing methods predominantly target hallucinations arising from the language component, leaving vision-induced hallucinations comparatively underexplored. In this work, we instead focus on hallucinations rooted in the visual modality by generating counterfactual images to extract and subsequently remove hallucination directions.

To this end, we propose CIPHER (Counterfactual Image Perturbations for Hallucination Extraction and Removal), a training-free, test-time hallucination mitigation method that performs feature-level intervention to address hallucinations in LVLMs. In the offline phase, we construct OHC-25K (Object Hallucination Counterfactuals), a 25K-sample diffusion-edited counterfactual dataset created from a subset of image–caption pairs from MSCOCO \cite{mscoco}. Figure~\ref{fig:motivation} illustrates how counterfactual images are generated from ground-truth image–caption pairs and incorporated into our framework. We first perturb each caption using the GPT-3.5 model \cite{gpt}, then apply a partial forward diffusion process to the image, and run the reverse diffusion process conditioned on the altered caption to synthesize a hallucinated variant. This controlled editing preserves the global structure of the original image while injecting semantically incorrect yet visually plausible elements (e.g., adding a cluster of grapes that are not present in the scene). We feed both the ground-truth and diffusion-edited image–caption pairs through the LVLM to extract hidden representations, compute their difference vectors, and stack them into a matrix. Applying Singular Value Decomposition (SVD) to these vectors reveals a low-rank subspace capturing the dominant directions of vision-induced hallucinations. At inference time, CIPHER suppresses hallucinations by projecting intermediate hidden representations away from this subspace, without modifying model weights or requiring any additional training.

We conduct extensive experiments across multiple benchmarks to evaluate our approach. Results demonstrate that it effectively mitigates hallucinations without degrading the overall performance of LVLMs on tasks such as image captioning and visual question answering. Unlike many existing methods, our solution operates entirely at test time and introduces no inference-time overhead, as hallucination suppression is achieved via a lightweight projection on intermediate hidden states. To the best of our knowledge, this is the first method to identify hallucination directions in the feature space by hallucinating images, enabling more targeted mitigation of vision-induced hallucinations. Our main contributions are as follows:

\textbf{(1)} We present \textbf{CIPHER}, a training-free test-time hallucination mitigation method that specifically targets hallucinations arising from the visual modality.
\textbf{(2)} We develop a diffusion-guided procedure to construct \textbf{OHC-25K}, a counterfactual dataset of diffusion-edited images whose semantics intentionally contradict their ground-truth captions. These counterfactual pairs enable us to estimate a hallucination subspace by contrasting them with clean image--caption representations.
\textbf{(3)} During inference, we mitigate hallucinations by projecting hidden representations onto the orthogonal complement of the estimated subspace, without requiring model retraining or parameter updates.
\textbf{(4)} Extensive experiments across multiple benchmarks demonstrate that {CIPHER} reduces hallucinations while preserving the overall quality of multimodal generation.

\section{Related Work}
\label{sec:related}
\vspace{-0.05cm}
\subsection{Large Vision-Language Models (LVLMs)}

The success of large language models (LLMs) \cite{brown2020language, chowdhery2023palm, llama2} has paved the way for large vision-language models (LVLMs), which extend language understanding to visual inputs. By integrating vision encoders with LLMs via adaptation modules and instruction tuning, LVLMs such as mPLUG-Owl2 \cite{ye2024mplug2}, Flamingo \cite{alayrac2022flamingo}, BLIP-2 \cite{li2023blip}, LLaVA \cite{LLaVA}, and MiniGPT-4 \cite{zhu2023minigpt} achieve strong performance on tasks like image captioning, visual question answering, and multimodal dialogue. However, they remain susceptible to hallucination, limiting robustness in real-world settings.

\vspace{-0.1cm}
\subsection{Hallucination Mitigation in LVLMs}

\begin{figure*}[!t]
    \centering
    \includegraphics[width=\linewidth]{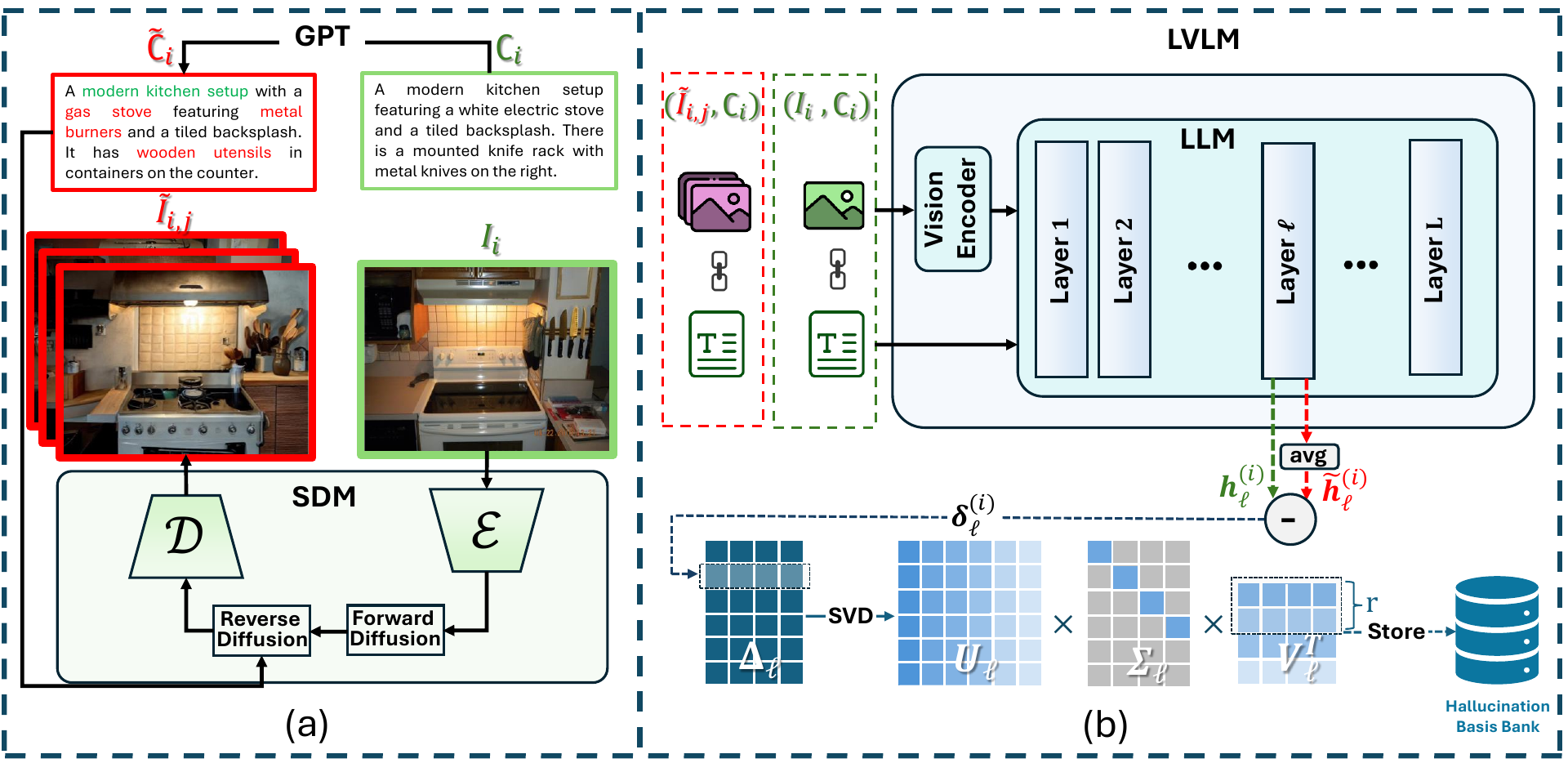}
\caption{
\textbf{(a)} Hallucinated image generation: given an image \( \boldsymbol{I}_i \) and its ground-truth caption \(\mathcal{C}_i\), a GPT model generates a hallucinated caption \(\tilde{\mathcal{C}}_i\). The image is then encoded by the encoder \( \mathcal{E} \) of the Stable Diffusion Model (SDM), and both forward and reverse diffusion steps are applied, conditioned on \(\tilde{\mathcal{C}}_i\), to produce  hallucinated images \( \tilde{\boldsymbol{I}}_{i,j} \). 
\textbf{(b)} Estimating hallucination subspace: the LVLM encodes both hallucinated and ground-truth image--caption pairs to extract hidden states. Feature differences \( \{ \boldsymbol{\delta}_\ell^{(i)} \}_{i=1}^M \) are computed, stacked, and decomposed via SVD. The top \( r \) right-singular vectors are retained in a hallucination basis bank for inference-time suppression.
}
    \label{fig:blockdiagram}
    \vspace{-0.3cm}
\end{figure*}

To mitigate hallucinations in LVLMs, various approaches have been proposed across the training, inference, and post-processing stages. While training-based methods~\cite{gunjal2024detecting, liu2023mitigating, jiang2024hallucination, op-dpo, bai2025mitigating, xiao2025detecting} are often effective, they typically require expensive annotations, model retraining, or architectural modifications, limiting their scalability in practice. In contrast, post-processing strategies mitigate hallucinations without altering the base model, typically by detecting and revising errors after generation using heuristics or external tools \cite{zhou2023lure, yin2024woodpecker}. However, dependence on auxiliary models limits generalizability and complicates deployment. Test-time approaches \cite{chen2024halc, liu2025reducing-vti, SID} address these limitations by intervening directly during inference, offering a more efficient solution.

\vspace{-0.05cm}
\subsection{Test-time Hallucination Suppression in LVLMs}
Test-time hallucination mitigation methods intervene during inference to improve factual alignment without modifying the model parameters. These include decoding-based techniques \cite{huang2023opera, leng2024mitigating-vcd, SID, an2025mitigating, yang2025mitigating, zhu2025ibd}, which steer generation by adjusting sampling distributions or applying constraints to suppress hallucinations. While effective, these approaches often involve multiple forward passes, leading to increased inference-time overhead. In contrast, feature-level interventions provide a more efficient solution. These methods operate by modifying internal representations within a single forward pass, guiding the model away from hallucinated outputs \cite{ITI, liu2025reducing-vti, yang2025nullu, uppaal2024model, chen2025ict}. For example, Nullu \cite{yang2025nullu} extract hallucination directions by contrasting features from textually perturbed inputs and shift the model’s hidden states accordingly. However, these approaches mostly focus on language-induced hallucinations, overlooking those rooted in the visual modality. CIPHER fills this gap by perturbing images with a diffusion model to extract the hallucination subspace and suppress it via feature projection.

\section{Method}

CIPHER estimates feature directions triggered by hallucination-inducing visual cues and suppresses these components during inference. The method operates in two stages: (1) an \textit{offline phase}, where we construct a counterfactual dataset and estimate the hallucination subspaces; and (2) an \textit{inference phase}, where we nullify hallucination-prone components in hidden representations during generation.
\subsection{Offline Phase}

\noindent \textbf{Counterfactual Dataset Generation (OHC-25K).}
To reveal hallucination-sensitive directions in the LVLM’s feature space, we construct a dataset of \textit{counterfactual image–caption pairs} in which the visual content is structurally preserved but semantically altered.

Figure~\ref{fig:blockdiagram}(a) illustrates how the counterfactual images are generated. We begin with \(M = 5{,}000\) randomly selected image--caption pairs from the MSCOCO training set,
\(\{(\boldsymbol{I}_i, \mathcal{C}_i)\}_{i=1}^{M}\).
Following LURE~\cite{zhou2023lure}, we generate a hallucinated caption \(\tilde{\mathcal{C}}_i\) (displayed with a red border) for each ground-truth caption \(\mathcal{C}_i\) (displayed with a green border) using a GPT-based perturbation that injects plausible but incorrect objects. 

Next, each image \(\boldsymbol{I}_i\) is encoded into the latent space of Stable Diffusion Model (SDM) ~\cite{rombach2022high} using its VAE encoder:

\begin{equation}
    \boldsymbol{z}_0 = \mathcal{E}(\boldsymbol{I}_i).
\end{equation}

We then apply \(t_h\) steps of the forward diffusion process to introduce Gaussian noise. 
Let \(\bar{\alpha}_{t_h}\) denote the cumulative product of noise scheduling coefficients up to step \(t_h\). 
The partially noised latent is given by:
\begin{equation}
    \tilde{\boldsymbol{z}}_{t_h}
    = \sqrt{\bar{\alpha}_{t_h}} \boldsymbol{z}_0
    + \sqrt{1 - \bar{\alpha}_{t_h}}\, \boldsymbol{\epsilon},
    \qquad
    \boldsymbol{\epsilon} \sim \mathcal{N}(0, I).
    \label{eq:forward-diff}
\end{equation}

To simulate semantic hallucination, we reverse-denoise $\tilde{\boldsymbol{z}}_{t_h}$ while conditioning on the hallucinated caption:
\begin{equation}
    \tilde{\boldsymbol{z}}_{t-1}
    = f_\theta(\tilde{\boldsymbol{z}}_t, t, \tilde{\mathcal{C}}_i),
    \qquad t = t_h, \dots, 1.
    \label{eq:reverse-diff}
\end{equation}

Decoding the final denoised latent (\(\tilde{\boldsymbol{z}}_0\)) yields a counterfactual image:
\begin{equation}
    \tilde{\boldsymbol{I}}_{i,j}
    = \mathcal{D}(\tilde{\boldsymbol{z}}_0),
\end{equation}
where \(j = 1, \dots, B\) indexes the \(B\) counterfactual variants generated from different Gaussian noise seeds (we use \(B = 5\)). Pairing each counterfactual image with the original ground-truth caption creates a semantic conflict while preserving structural alignment.  
This forms the \textit{Object-Hallucination Counterfactuals} dataset, constructed from \(M = 5{,}000\) image--caption pairs with \(B = 5\) counterfactual variants per image, resulting in \(25{,}000\) images.

\begin{equation}
    \textbf{OHC-25K}
    =
    \{(\tilde{\boldsymbol{I}}_{i,j}, \mathcal{C}_i)
    \;|\;
    i = 1,\dots,M,\;
    j = 1,\dots,B\}.
\end{equation}

\vspace{0.1cm}
\noindent \textbf{Estimating the Hallucination Subspace.}
Figure \ref{fig:blockdiagram}(b) illustrates how the hallucination subspace is extracted. For each original pair $(\boldsymbol{I}_i, \mathcal{C}_i)$ and its $B$ counterfactual variants
$(\tilde{\boldsymbol{I}}_{i,j}, \mathcal{C}_i)$, we extract hidden representations from the frozen LVLM. Let $\boldsymbol{h}_{\ell,k}^{(i)} \in \mathbb{R}^d$ denote the hidden state of the $k$-th caption token at layer $\ell$ for the original pair, and
$\tilde{\boldsymbol{h}}_{\ell,k}^{(i,j)} \in \mathbb{R}^d$ for the $j$-th counterfactual variant. We mean-pool across caption tokens:
\begin{equation}
    \boldsymbol{h}_{\ell}^{(i)}
    = \frac{1}{N} \sum_{k=1}^N \boldsymbol{h}_{\ell,k}^{(i)},
    \qquad
    \tilde{\boldsymbol{h}}_{\ell}^{(i,j)}
    = \frac{1}{N} \sum_{k=1}^N \tilde{\boldsymbol{h}}_{\ell,k}^{(i,j)}.
\end{equation}

Since each image yields $B$ hallucinated samples, we compute the aggregated counterfactual representation as:
\begin{equation}
    \tilde{\boldsymbol{h}}_{\ell}^{(i)}
    = \frac{1}{B} \sum_{j=1}^{B}
    \tilde{\boldsymbol{h}}_{\ell}^{(i,j)}.
\end{equation}

The hallucination direction for sample \(i\) at layer \(\ell\) is defined as 
\(\boldsymbol{\delta}_{\ell}^{(i)} = \tilde{\boldsymbol{h}}_{\ell}^{(i)} - \boldsymbol{h}_{\ell}^{(i)}\).
Stacking these directions across all samples yields:

\begin{equation}
    \boldsymbol{\Delta}_\ell
    =
    \begin{bmatrix}
        \boldsymbol{\delta}_{\ell}^{(1)};
        \boldsymbol{\delta}_{\ell}^{(2)};
        \cdots;
        \boldsymbol{\delta}_{\ell}^{(M)}
    \end{bmatrix}
    \in \mathbb{R}^{M \times d}.
\end{equation}

We apply Singular Value Decomposition (SVD) to the stacked difference matrix to identify the principal hallucination directions:
\begin{equation}
    \boldsymbol{\Delta}_\ell
    = \boldsymbol{U}_\ell
    \boldsymbol{\Sigma}_\ell
    \boldsymbol{V}_\ell^\top.
\end{equation}

The top $r$ right-singular vectors \(
    \boldsymbol{V}_{\ell,r}
    =
    [\boldsymbol{v}_{\ell,1}, \dots, \boldsymbol{v}_{\ell,r}]
\)
span the hallucination subspace at layer $\ell$.  
These vectors form our \textit{hallucination basis bank}.

\begin{figure}
    \centering
    \includegraphics[width=\linewidth]{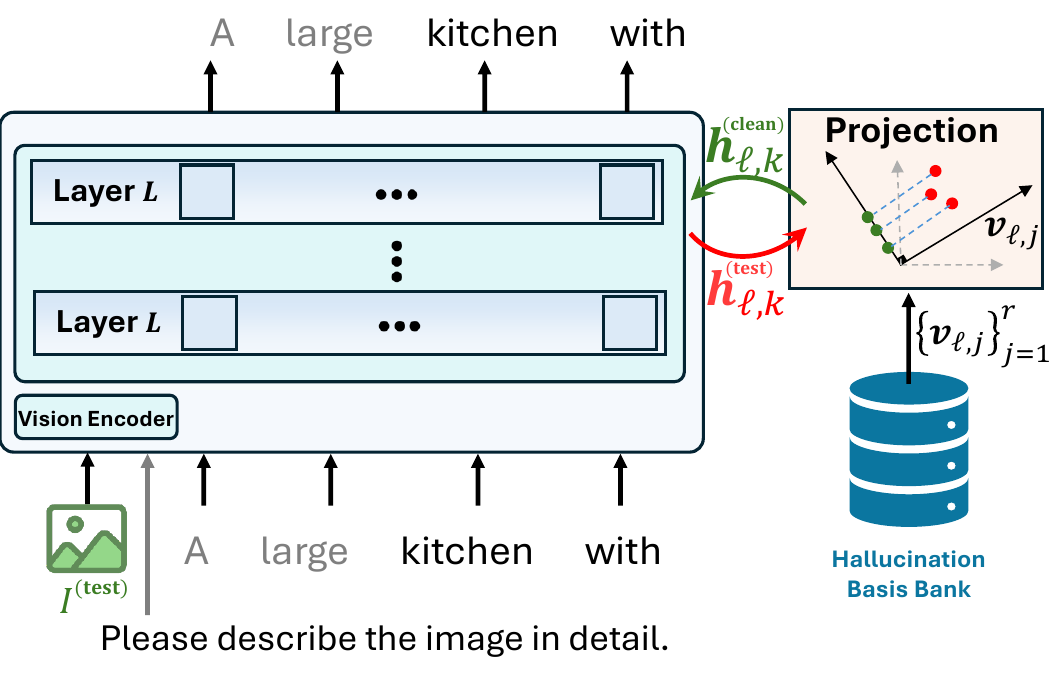}
    \caption{Inference-time projection mechanism. Given an input image and instruction, the model generates text autoregressively. At each decoding step during generation, hidden states from selected layers are projected onto the subspace orthogonal to the corresponding hallucination space, using the hallucination basis bank obtained in the offline phase.}
    \label{fig:inference}
\end{figure}

\subsection{Inference Phase}

\vspace{0.1cm}
\noindent \textbf{Test-Time Nullification.}
The hallucination nullification process during test time is illustrated in Figure \ref{fig:inference}.  
At each decoding step \(k\) and selected layer \(\ell\), the hidden state of the test image during inference is denoted by \(\boldsymbol{h}_{\ell,k}^{\text{test}} \in \mathbb{R}^d\). To suppress hallucinations, we remove components aligned with the hallucination subspace:
\begin{equation}
    \boldsymbol{h}_{\ell,k}^{\text{clean}}
    =
    \boldsymbol{h}_{\ell,k}^{\text{test}}
    -
    \sum_{j=1}^{r}
    \langle
        \boldsymbol{h}_{\ell,k}^{\text{test}},
        \boldsymbol{v}_{\ell,j}
    \rangle
    \boldsymbol{v}_{\ell,j}.
\end{equation}

Equivalently, this operation can be expressed using the projection matrix
\(
    \boldsymbol{P}_\ell
    = \boldsymbol{I}
    - \boldsymbol{V}_{\ell,r} \boldsymbol{V}_{\ell,r}^\top,
\)
which yields:
\begin{equation}
    \boldsymbol{h}_{\ell,k}^{\text{clean}}
    = \boldsymbol{P}_\ell \, \boldsymbol{h}_{\ell,k}^{\text{test}}.
\end{equation}

This layer-wise nullification is applied before decoding each token, removing hallucination-prone directions while preserving core semantics.

\section{Experiments}
\label{sec:experiments}

\begin{table*}[!t]
\centering
\renewcommand{\arraystretch}{1.25}
\resizebox{\textwidth}{!}{%
\begin{tabular}{l|ccc|ccc|ccc}
\toprule
\rowcolor{headergray}
\textbf{Method} & \multicolumn{3}{c|}{\textbf{LLaVA-1.5}} & \multicolumn{3}{c|}{\textbf{MiniGPT-4}} & \multicolumn{3}{c}{\textbf{mPLUG-Owl2}} \\
\rowcolor{headergray}
 & \textbf{CHAIR$_S \downarrow$} & \textbf{CHAIR$_I \downarrow$} & \textbf{BLEU$\uparrow$}
 & \textbf{CHAIR$_S \downarrow$} & \textbf{CHAIR$_I \downarrow$} & \textbf{BLEU$\uparrow$}
 & \textbf{CHAIR$_S \downarrow$} & \textbf{CHAIR$_I \downarrow$} & {BLEU$\uparrow$} \\
\midrule
\textbf{Greedy} & 20.40$_{\pm2.80}$ & 7.08$_{\pm0.33}$ & 15.72$_{\pm0.10}$ 
       & 32.40$_{\pm2.20}$ & 12.20$_{\pm0.42}$ & 14.57$_{\pm0.11}$ 
       & 22.90$_{\pm0.90}$ & 8.62$_{\pm0.11}$ & 15.01$_{\pm0.24}$ \\
\textbf{Beam Search} 
       & 19.50$_{\pm2.30}$ & 6.84$_{\pm0.79}$ & 15.99$_{\pm0.14}$ 
       & 30.10$_{\pm0.30}$ & 11.87$_{\pm0.37}$ & 15.35$_{\pm0.24}$ 
       & 20.30$_{\pm0.70}$ & 7.62$_{\pm0.19}$ & 15.43$_{\pm0.05}$ \\
\textbf{DoLa} \small{(ICLR'24)}
       & 20.20$_{\pm2.80}$ & 6.75$_{\pm0.54}$ & 15.68$_{\pm0.10}$ 
       & 31.90$_{\pm3.30}$ & 12.15$_{\pm0.89}$ & 14.54$_{\pm0.12}$ 
       & 22.40$_{\pm1.80}$ & 8.36$_{\pm0.04}$ & 15.13$_{\pm0.21}$ \\
\textbf{OPERA} \small{(CVPR'24)}
       & 17.50$_{\pm0.50}$ & 6.07$_{\pm0.32}$ & 16.02$_{\pm0.02}$ 
       & 29.70$_{\pm0.30}$ & 11.96$_{\pm0.29}$ & 14.82$_{\pm0.05}$ 
       & 20.07$_{\pm2.07}$ & 7.18$_{\pm0.39}$ & 15.41$_{\pm0.12}$ \\
\textbf{VCD} \small{(CVPR'24)}
       & 20.30$_{\pm1.10}$ & 7.28$_{\pm0.10}$ & 14.53$_{\pm0.01}$ 
       & 29.00$_{\pm2.80}$ & 12.64$_{\pm1.19}$ & 14.42$_{\pm0.01}$ 
       & 22.80$_{\pm0.80}$ & 8.68$_{\pm0.17}$ & 15.14$_{\pm0.13}$ \\
\textbf{Woodpecker} \small{(SCIS'24)}
       & 23.85$_{\pm4.62}$ & 7.50$_{\pm0.01}$ & 17.05$_{\pm0.00}$ 
       & 28.87$_{\pm2.20}$ & 10.20$_{\pm0.85}$ & 15.30$_{\pm0.01}$ 
       & 26.33$_{\pm1.98}$ & 8.43$_{\pm0.80}$ & 16.43$_{\pm0.00}$ \\
\textbf{LURE} \small{(ICLR'24)}
       & 19.48$_{\pm2.35}$ & 6.50$_{\pm0.38}$ & 15.97$_{\pm0.01}$ 
       & 27.88$_{\pm2.25}$ & 10.20$_{\pm0.85}$ & 15.03$_{\pm0.01}$ 
       & 21.27$_{\pm0.06}$ & 7.67$_{\pm0.16}$ & 15.65$_{\pm0.15}$ \\
\textbf{HALC} \small{(ICML'24)}
       & 16.90$_{\pm2.10}$ & 5.72$_{\pm0.55}$ & 16.02$_{\pm0.04}$ 
       & 25.20$_{\pm2.00}$ & 9.42$_{\pm0.41}$ & 14.91$_{\pm0.13}$ 
       & 18.80$_{\pm1.20}$ & 7.00$_{\pm0.01}$ & 15.33$_{\pm0.24}$ \\
\textbf{Nullu} \small{{(CVPR'25)}} 
       & {15.20$_{\pm0.60}$} & {5.30$_{\pm0.03}$} & {15.69$_{\pm0.04}$} 
       & {21.40$_{\pm1.00}$} & {8.99$_{\pm0.36}$} & {14.81$_{\pm0.06}$} 
       & {15.60$_{\pm1.20}$} & {5.77$_{\pm0.01}$} & {15.45$_{\pm0.01}$} \\
\midrule
\textbf{CIPHER (Ours)} & \textbf{13.05}$_{\pm 0.57}$  & \textbf{4.53}$_{\pm0.38}$ & 15.82$_{\pm0.25}$ 
              & \textbf{18.48}$_{\pm 1.20}$ & \textbf{8.33}$_{\pm 0.17}$ & 15.10$_{\pm0.43}$
              & \textbf{13.60}$_{\pm1.06}$ & \textbf{4.92}$_{\pm0.15}$ & 16.25$_{\pm0.47}$ \\
\bottomrule
\end{tabular}
}
\caption{CHAIR and BLEU scores across LVLMs; lower CHAIR = less hallucination, higher BLEU = better fluency.}
\label{tab:results-chair}
\end{table*}

\subsection{Implementation Details}
We evaluate CIPHER on CHAIR \cite{chair} and Offline POPE (OPOPE) \cite{chen2024halc} using LLaVA-1.5~\cite{LLaVA} with Vicuna~\cite{chiang2023vicuna}, MiniGPT-4~\cite{zhu2023minigpt} with LLaMA2~\cite{llama2}, and mPLUG-Owl2~\cite{ye2024mplug2}. For decoding, we use a beam size of 3, setting the maximum number of generated tokens to 64 for CHAIR and 256 for OPOPE. The hallucination subspace is constructed using the top \( r \) singular vectors, with \( r = 8 \) for LLaVA, 64 for MiniGPT-4, and 32 for mPLUG-Owl2, selected via grid search. Following \cite{yang2025nullu}, we apply the projection at upper layers (16--32). Visual hallucinations are generated using Stable Diffusion v1.5, with noise added up to \(t_h = 0.5T\), where \(T\) denotes the total number of diffusion steps. We use a classifier-free guidance scale of 7.5. See the Appendix for additional details.

\subsection{Datasets and Benchmarks}
For the offline phase, we follow LURE~\cite{zhou2023lure} using 5,000 image-caption pairs from the MSCOCO training split \cite{mscoco}. For evaluation, we follow established evaluation protocols~\cite{huang2023opera, yin2024woodpecker} and randomly sample 500 images from the MSCOCO validation set using the widely adopted CHAIR and POPE metrics. We report mean and standard deviation over three runs with different random seeds.

\vspace{0.1cm}
\noindent \textbf {CHAIR.} We use the CHAIR benchmark \cite{chair} to evaluate object hallucination in generated image descriptions. CHAIR$_S$ measures the percentage of sentences that contain at least one hallucinated object, while CHAIR$_I$ captures the proportion of hallucinated objects among all generated object mentions. We also report BLEU \cite{papineni2002bleu} to evaluate text quality and confirm that fluency is preserved.

\vspace{0.1cm}
\noindent \textbf{OPOPE.} We evaluate object hallucination using the Offline POPE \cite{pope}, which includes three negative sampling strategies: random, popular, and adversarial. We adopt the OPOPE protocol \cite{chen2024halc}, which applies a CHAIR-style captioning setup and modifies the evaluation metric to detect the inclusion of negative objects in generated captions.

\vspace{0.1cm}
\noindent \textbf{ MMHAL-Bench.}
MMHal \cite{MMHAL} evaluates LVLMs across eight hallucination types, including attributes, adversarial objects, comparisons, counting, relations, environment, holistic descriptions, and others, providing a broad assessment beyond object-level errors. We measure the response informativeness using GPT-4.

\vspace{0.1cm}
\noindent \textbf{LLaVA-Bench.} We also evaluate on LLaVA-Bench \cite{LLaVA2}, a curated benchmark of 24 diverse images with human-written descriptions and targeted questions. It assesses accuracy and detailedness, complementing CHAIR and POPE. Following \cite{leng2024mitigating-vcd, yin2024woodpecker}, we use GPT-4V to compare original and edited responses based on their alignment with the image and question.

\subsection{Baselines}

We compare CIPHER with state-of-the-art hallucination mitigation methods, categorized as test-time approaches and post-processing strategies. Test-time methods include DoLa \cite{chuang2023dola}, VCD \cite{leng2024mitigating-vcd}, HALC \cite{chen2024halc}, OPERA \cite{huang2023opera}, and Nullu \cite{yang2025nullu}. The first four apply decoding-based interventions, while Nullu projects hidden representations onto a hallucination-free subspace. Post-processing methods Woodpecker \cite{yin2024woodpecker} and LURE \cite{zhou2023lure} detect and rewrite hallucinated spans using external models.

\begin{table*}[!t]
\centering
\resizebox{\textwidth}{!}{%
\setlength{\tabcolsep}{4pt}
\renewcommand{\arraystretch}{1.3}
\begin{tabular}{l|ccc|ccc|ccc}
\toprule
\rowcolor{headergray}
\textbf{Method} & \multicolumn{3}{c|}{\textbf{LLaVA-1.5}} & \multicolumn{3}{c|}{\textbf{MiniGPT-4}} & \multicolumn{3}{c}{\textbf{mPLUG-Owl2}} \\
\rowcolor{headergray}
& \textbf{Accuracy$\uparrow$} & \textbf{Precision$\uparrow$} & \textbf{F score$\uparrow$} & \textbf{Accuracy$\uparrow$} & \textbf{Precision$\uparrow$} & \textbf{F score$\uparrow$} & \textbf{Accuracy$\uparrow$} & \textbf{Precision$\uparrow$} & \textbf{F score$\uparrow$} \\
\midrule
\textbf{Greedy} & 79.14{\tiny$\pm$0.89} & 91.98{\tiny$\pm$0.82} & 90.45{\tiny$\pm$0.86} & 71.22{\tiny$\pm$1.27} & 93.72{\tiny$\pm$1.02} & 90.04{\tiny$\pm$1.23} & 76.46{\tiny$\pm$0.92} & 88.85{\tiny$\pm$1.15} & 87.29{\tiny$\pm$1.15} \\
\textbf{Beam Search} & 79.41{\tiny$\pm$0.69} & 92.52{\tiny$\pm$0.55} & 90.96{\tiny$\pm$0.59} & 71.65{\tiny$\pm$1.15} & 94.70{\tiny$\pm$0.60} & 90.97{\tiny$\pm$0.85} & 76.76{\tiny$\pm$1.02} & 90.28{\tiny$\pm$0.80} & 88.56{\tiny$\pm$0.87} \\
\textbf{DoLa} \small{(ICLR'24)} & 78.98{\tiny$\pm$0.56} & 91.66{\tiny$\pm$0.81} & 90.15{\tiny$\pm$0.79} & 71.28{\tiny$\pm$1.15} & 93.92{\tiny$\pm$0.83} & 90.22{\tiny$\pm$1.04} & 76.07{\tiny$\pm$1.09} & 88.54{\tiny$\pm$1.25} & 86.95{\tiny$\pm$1.27} \\
\textbf{OPERA} \small{(CVPR'24)} & 79.29{\tiny$\pm$0.32} & 92.25{\tiny$\pm$0.07} & 90.71{\tiny$\pm$0.11} & 70.48{\tiny$\pm$1.63} & 94.41{\tiny$\pm$1.11} & 90.66{\tiny$\pm$1.42} & 75.49{\tiny$\pm$1.29} & 91.23{\tiny$\pm$1.06} & 89.11{\tiny$\pm$1.17} \\
\textbf{VCD} \small{(CVPR'24)} & 78.01{\tiny$\pm$0.75} & 91.33{\tiny$\pm$0.88} & 89.69{\tiny$\pm$0.89} & 70.83{\tiny$\pm$1.83} & 92.31{\tiny$\pm$0.88} & 88.76{\tiny$\pm$1.29} & 75.49{\tiny$\pm$1.27} & 88.75{\tiny$\pm$1.56} & 87.02{\tiny$\pm$1.57} \\
\textbf{HALC} \small{(ICML'24)} & 77.87{\tiny$\pm$0.22} & 93.17{\tiny$\pm$0.39} & 91.25{\tiny$\pm$0.38} & 71.17{\tiny$\pm$0.89} & 94.88{\tiny$\pm$0.15} & 90.95{\tiny$\pm$0.42} & 74.93{\tiny$\pm$1.09} & 90.20{\tiny$\pm$0.90} & 88.12{\tiny$\pm$0.99} \\
\textbf{Nullu} \small{(CVPR'25)} & 79.52{\tiny$\pm$0.04} & 93.46{\tiny$\pm$0.03} & 91.79{\tiny$\pm$0.04} & 71.92{\tiny$\pm$0.39} & 95.96{\tiny$\pm$0.65} & 92.07{\tiny$\pm$0.65} & 77.09{\tiny$\pm$1.37} & 92.83{\tiny$\pm$0.29} & 90.80{\tiny$\pm$0.52} \\
\midrule
\textbf{CIPHER (Ours)} & \textbf{80.05}{\tiny$\pm$0.07}& \textbf{93.72}{\tiny$\pm$0.14} & \textbf{92.11}{\tiny$\pm$0.11} & \textbf{72.25}{\tiny$\pm$0.26} & \textbf{96.50}{\tiny$\pm$0.19} & \textbf{92.58}{\tiny$\pm$0.20} & \textbf{77.87}{\tiny$\pm$0.21} & \textbf{92.93}{\tiny$\pm$0.26} & \textbf{90.95}{\tiny$\pm$0.59} \\
\bottomrule
\end{tabular}%
}
\caption{Accuracy, precision, and F1 scores of different methods on the OPOPE benchmark, evaluated across LLaVA-1.5, MiniGPT-4, and mPLUG-Owl2.}
\label{tab:opope}
\end{table*}

\subsection{Results on CHAIR}

We prompt all methods with the instruction: “Please describe this image in detail” and report the results in Table~\ref{tab:results-chair}. CIPHER consistently achieves the lowest hallucination rates across all three vision-language models (LLaVA-1.5, MiniGPT-4, and mPLUG-Owl2). For example, on LLaVA-1.5, CIPHER reduces CHAIR$_S$ to 13.05\%, which is a 2.15\% absolute improvement over the second-best model (Nullu at 15.20\%), and a 7.35\% drop compared to the original greedy baseline (20.40\%). On MiniGPT-4, CIPHER further brings CHAIR$_S$ down to 18.48\%, improving upon Nullu by 2.92\% and reducing hallucinations by a substantial 13.92\% relative to the greedy baseline (32.40\%). Similarly, for mPLUG-Owl2, CIPHER scores 13.60\% on CHAIR$_S$, which is 2.00\% better than the second-best model and 9.30\% better than the original model. These gains in CHAIR$_S$ are mirrored in CHAIR$_I$ as well, with CIPHER consistently outperforming all baselines in object-level hallucination. Importantly, CIPHER maintains or improves BLEU scores across all models, indicating effective hallucination mitigation without sacrificing generation richness.

\subsection{Results on OPOPE}

As shown in Table~\ref{tab:opope}, CIPHER achieves the highest scores across all three evaluation metrics for each model, clearly demonstrating its effectiveness in mitigating object hallucinations while preserving answer quality. The gains in precision are particularly pronounced relative to the unedited model, further highlighting CIPHER’s ability to suppress hallucinations while retaining relevant content. Despite the already saturated performance regime, CIPHER continues to deliver consistent improvements. Detailed POPE and OPOPE results are provided in the Appendix.

\begin{figure}
    \centering
    \includegraphics[width=0.68\linewidth]{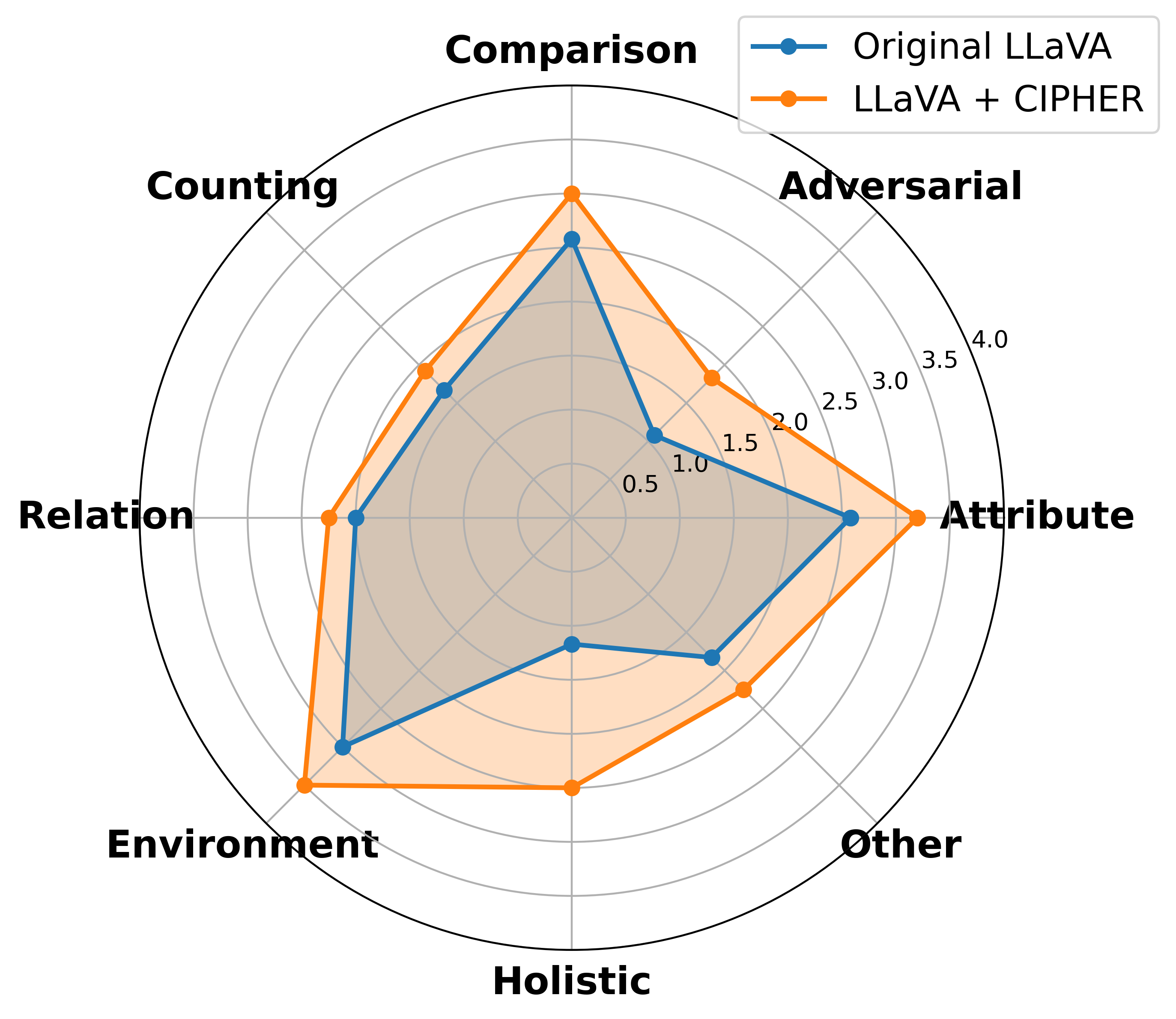}
    \vspace{-0.2cm}
    \caption{Radar chart of MMHal scores.}
    \label{fig:mmhal}
\end{figure}

\subsection{MMHal Benchmark Results}

We evaluate hallucination reduction using the MMHal benchmark, which covers eight categories. This benchmark provides a fine-grained view of different hallucination types. As shown in Fig.~\ref{fig:mmhal}, our {LLaVA + CIPHER} model achieves consistent improvements across all categories compared to the original LLaVA. The largest gains occur in \textit{Attribute}, \textit{Environment}, \textit{Holistic} and \textit{Adversarial} cases, while improvements in \textit{Relation} and \textit{Comparison} demonstrate enhanced global reasoning and grounding.


\begin{table}[!t]
\centering
\small
\renewcommand{\arraystretch}{1.1}
\setlength{\tabcolsep}{7.5pt}
\begin{tabular}{llcc}
\rowcolor{lightgray}
\toprule
\textbf{Model} & \textbf{Method} & \textbf{Accuracy$\uparrow$} & \textbf{Detailedness$\uparrow$} \\
\midrule
\multirow{2}{*}{LLaVA-1.5} & Original & 6.79$_{\pm1.50}$ & 6.33$_{\pm1.18}$ \\
                           & \textbf{CIPHER} & \textbf{7.08}$_{\pm1.32}$ & \textbf{6.75}$_{\pm1.13}$ \\
\midrule
\multirow{2}{*}{MiniGPT-4} & Original & 5.54$_{\pm2.41}$ & 5.58$_{\pm2.06}$ \\
                           & \textbf{CIPHER} & \textbf{6.16}$_{\pm1.99}$ & \textbf{6.64}$_{\pm1.88}$ \\
\midrule
\multirow{2}{*}{mPLUG-Owl2} & Original & 6.48$_{\pm2.12}$ & 6.26$_{\pm1.62}$ \\
                            & \textbf{CIPHER} & \textbf{7.35}$_{\pm1.63}$ & \textbf{6.91}$_{\pm1.61}$ \\
\bottomrule
\end{tabular}
\caption{GPT-4V evaluations on LLaVA-Bench indicate that CIPHER consistently enhances model outputs, yielding higher accuracy and more detailed answers.}
\label{tab:llava-bench}
\end{table}

\begin{figure}[!t]
    \centering
    \includegraphics[width=\linewidth]{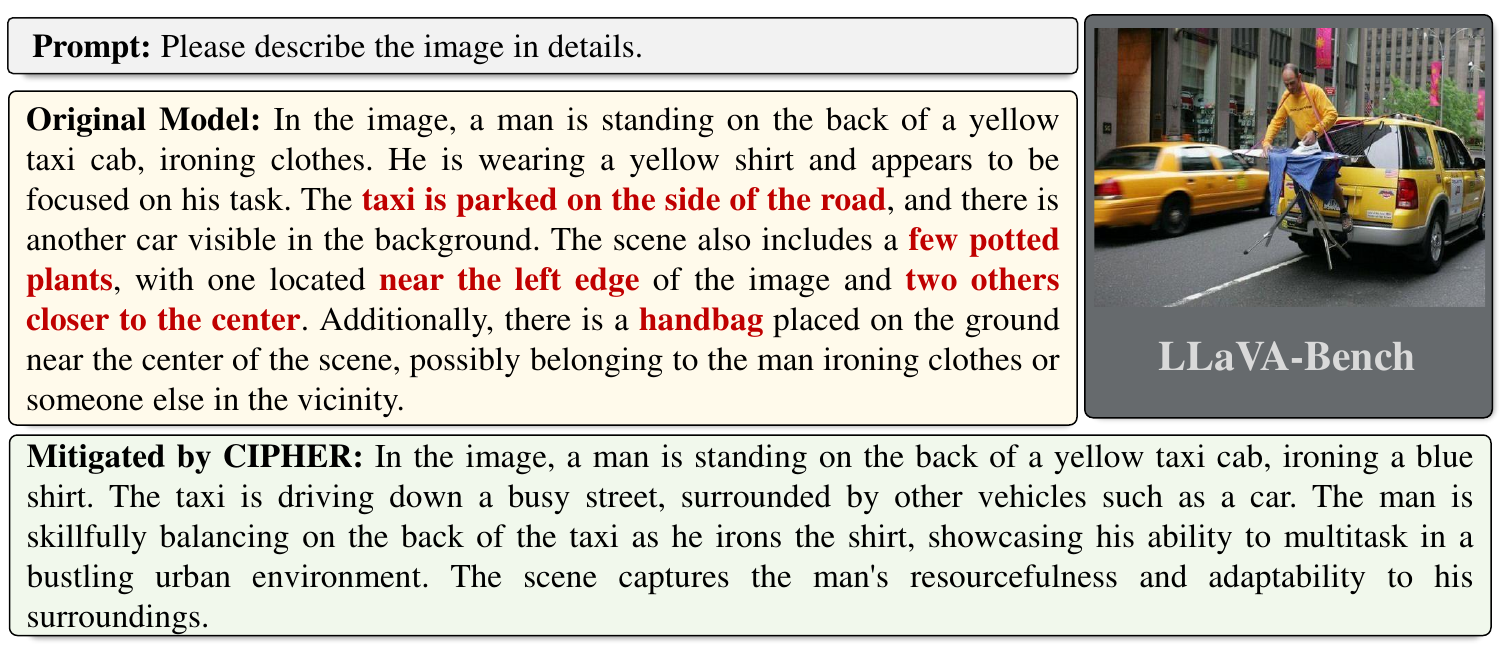}
    \caption{LLaVA-Bench example: CIPHER reduces hallucinations and improves grounding.}
    \label{fig:llava-bench-example}
        \vspace{-0.2cm}
\end{figure}

\begin{figure*}[!t]
    \centering
    \includegraphics[width=\linewidth]{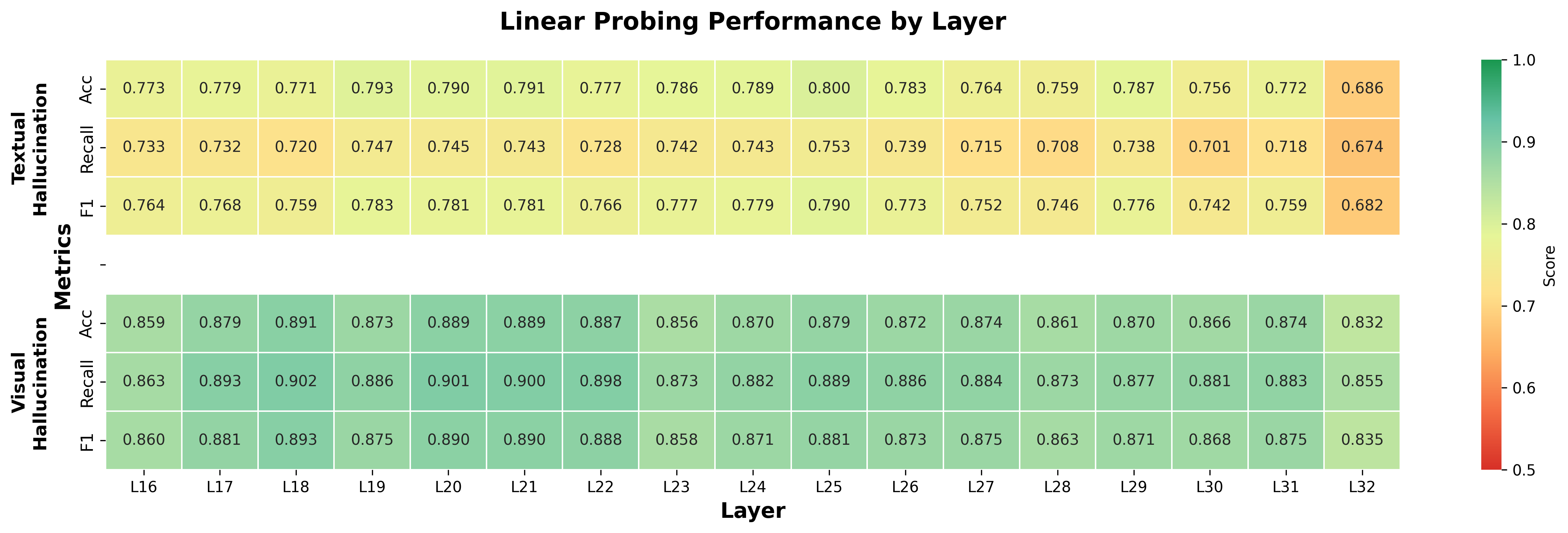}
    \caption{Layer-wise linear probing performance comparing textual and diffusion-based visual hallucination perturbations. Textual perturbations exhibit moderate and unstable separability across layers, whereas diffusion-based visual perturbations produce consistently high accuracy, recall, and F1.}
    \label{fig:linear-probing}
\end{figure*}
\subsection{Evaluation on LLaVA-Bench}

To further evaluate the effectiveness of our hallucination suppression strategy, we conduct quantitative and qualitative assessments on LLaVA-Bench~\cite{LLaVA2}. GPT-4V is prompted to assign scores from 1 to 10 for both \textit{accuracy} and \textit{detailedness}, based on the input image, question, and model response (see Appendix for the prompt). As shown in Table~\ref{tab:llava-bench}, applying CIPHER consistently improves both metrics across all three models, reflecting enhanced visual grounding and reduced hallucination. We also present a qualitative case study using LLaVA-1.5-7B in Figure~\ref{fig:llava-bench-example}. The original model generates a caption that includes several hallucinated elements, such as “the taxi is parked,” “potted plants,” and a “handbag,” none of which are visible in the image. In contrast, the CIPHER-edited description accurately captures the scene—a man ironing a shirt on the back of a moving yellow taxi amidst traffic—without introducing spurious objects.

\subsection{Inference Time}

We evaluate the efficiency of various hallucination mitigation methods in terms of throughput, measured in items per second using LLaVA-7B on an NVIDIA A6000 GPU (Table~\ref{tab:chair_throughput_compact}). CIPHER matches the throughput of standard greedy decoding ({0.70} items/s), substantially exceeding the efficiency of prior mitigation methods such as OPERA and HALC, which incur significant latency. In addition to its speed, CIPHER also achieves a much lower CHAIR$_S$ score ({13.05}\%), demonstrating superior hallucination suppression compared to all other baselines.

\begin{table}[!h]
\centering
\scriptsize
\setlength{\tabcolsep}{1.8pt}
\renewcommand{\arraystretch}{1.15}
\begin{tabular}{l|cccccccc}
\midrule
\rowcolor{lightgray}
\textbf{Metric} & Greedy & Beam & DoLa & VCD & OPERA & HALC & Nullu & CIPHER\\
\midrule
CHAIR$_S $ $\downarrow$ (\%) & 20.40 & 19.50 & 20.20 & 20.30 & 17.50 & 16.90 & 15.20 & \textbf{13.05} \\
Throughput $\uparrow$ (items/s) & \textbf{0.70} & 0.55 & 0.42 & 0.35 & 0.10 & 0.05 & \textbf{0.70} & \textbf{0.70} \\
\bottomrule
\end{tabular}
\caption{Comparison of CHAIR$_S$ and throughput (items/s) for different mitigation methods, tested on LLaVA-7B with an NVIDIA A6000 GPU.}
\label{tab:chair_throughput_compact}
\end{table}

\subsection{Ablation Studies}

\noindent \textbf{Linear Probing Analysis.}
To understand why the visual hallucination subspace is more effective than its text-based counterpart (Nullu method), we perform a layer-wise linear probing experiment. For each transformer layer, we train a logistic regression classifier to distinguish clean hidden representations from perturbed ones, using 400 samples for training and 1{,}000 samples for testing. We evaluate two perturbation types: (i) \textit{textual hallucination}, where the clean image is paired with a GPT-generated hallucinated caption; and (ii) \textit{visual hallucination}, where the caption remains unchanged but the image is perturbed using CIPHER-style diffusion counterfactuals.

As shown in Fig.~\ref{fig:linear-probing}, textual perturbations lead to only moderate separability (Accuracy: 0.73--0.80, F1: 0.74--0.78), with substantial variability across layers. This indicates that text-induced hallucination signals are relatively weak and noisy in the hidden representation space. In contrast, visual perturbations produce consistently high linear separability across all layers (Accuracy: 0.86--0.89, F1: 0.86--0.89). The diffusion counterfactuals introduce strong, structured, and layer-stable representation shifts that are easily captured by a linear classifier. These findings support our approach: the subspace derived from visual perturbations captures more meaningful hallucination directions, explaining CIPHER’s superior hallucination suppression.

\begin{figure}[!t]
    \centering
    \includegraphics[width=0.95\linewidth]{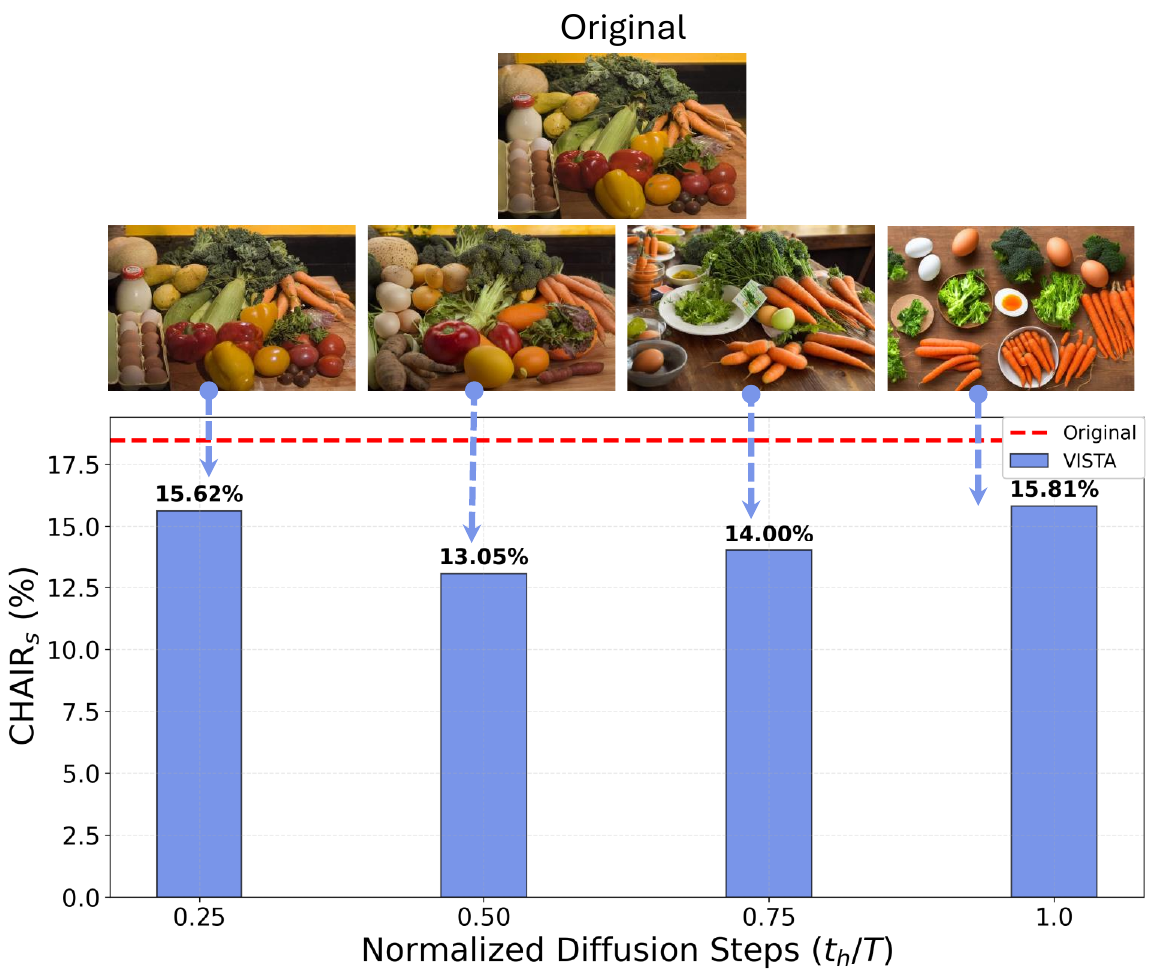}
\caption{CHAIR\(_S\) of CIPHER using hallucination subspaces derived from images perturbed at different diffusion steps.}

    \label{fig:ablation_diffusion_strength}
\end{figure}

\vspace{0.1cm}
\noindent \textbf{Effect of Diffusion Time Step.}
We investigate how the strength of semantic perturbation, controlled by the diffusion timestep (\(t_h\)) impacts the quality of the hallucination subspace in CIPHER. To assess the impact of diffusion strength, we evaluate four timesteps \( t_h \in \{0.25T, 0.5T, 0.75T, T\} \), where \( T \) is the maximum diffusion step. At each \( t_h \), we perform a round-trip diffusion: forward corruption and reverse denoising conditioned on a hallucinated caption, using Equation \ref{eq:forward-diff} and Equation \ref{eq:reverse-diff}, respectively. This produces images with increasing semantic shift. We compute hidden-state differences between hallucinated and original pairs, apply SVD to extract the hallucination subspace, and use it for test-time projection. As shown in Figure~\ref{fig:ablation_diffusion_strength}, the subspace from \( t_h = 0.5T \) achieves the best CHAIR\textsubscript{S}. We attribute this to intermediate steps yielding structurally faithful yet semantically altered images, striking a balance that best isolates hallucination directions.

\begin{figure}[!t]
    \centering
    \includegraphics[width=\linewidth]{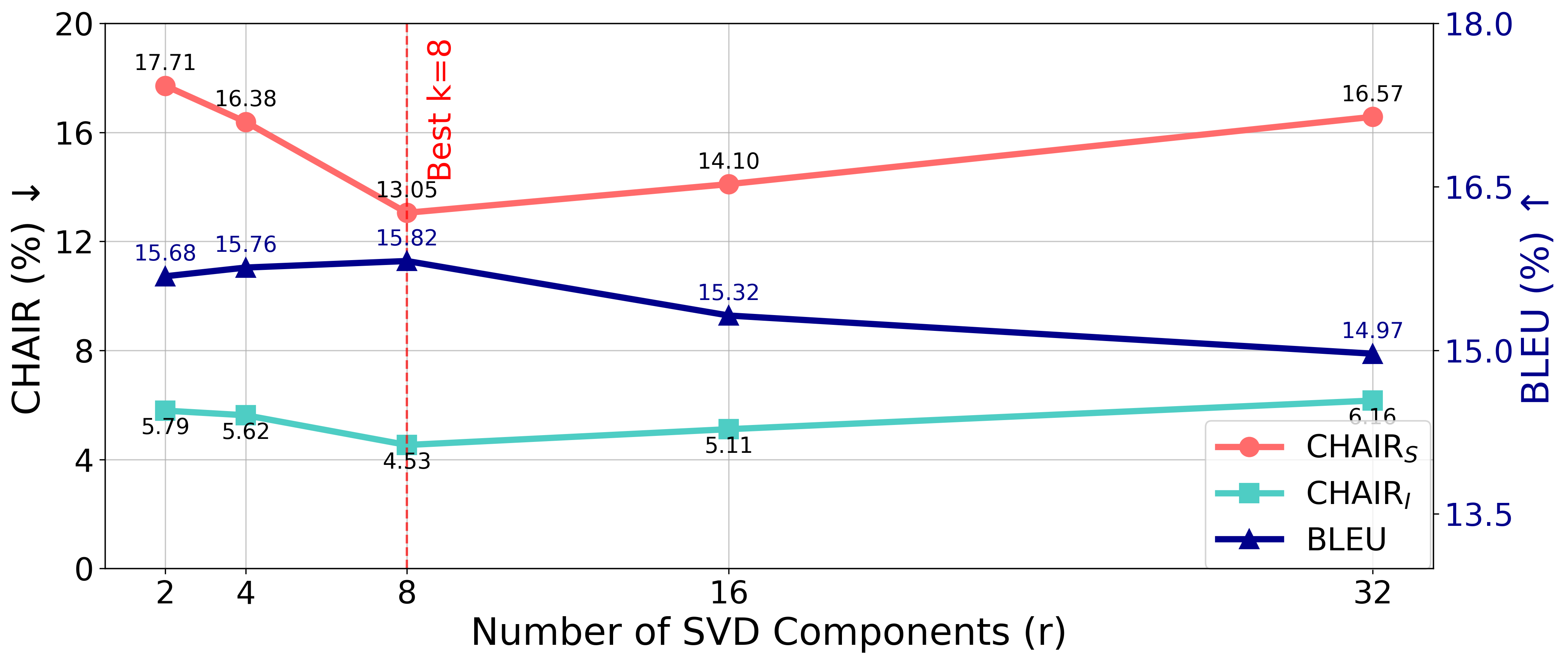}
    \vspace{-0.6cm}
    \caption{Ablation study on subspace rank (\(r\))}
    \label{fig:ablation_rank_llava}
\end{figure}

\vspace{0.1cm}
\noindent \textbf{Effect of Subspace Rank.}
\label{sec:svd_ablation}
We study how the number of singular vectors  (\( r \)) used to define the hallucination subspace affects performance. For the LLaVA-7B model, we sweep \( r \in \{2, 4, 8, 16, 32\} \) and evaluate on the CHAIR benchmark. As shown in Figure~\ref{fig:ablation_rank_llava}, \( r = 8 \) yields the best results: CHAIR$_{S}$ and CHAIR$_{I}$ are minimized, while BLEU reaches its peak, indicating improved faithfulness and fluency. Further component-selection results for MiniGPT-4 and mPLUG-Owl2 are given in the Appendix.

\begin{figure}[!t]
    \centering
    \includegraphics[width=0.95\linewidth]{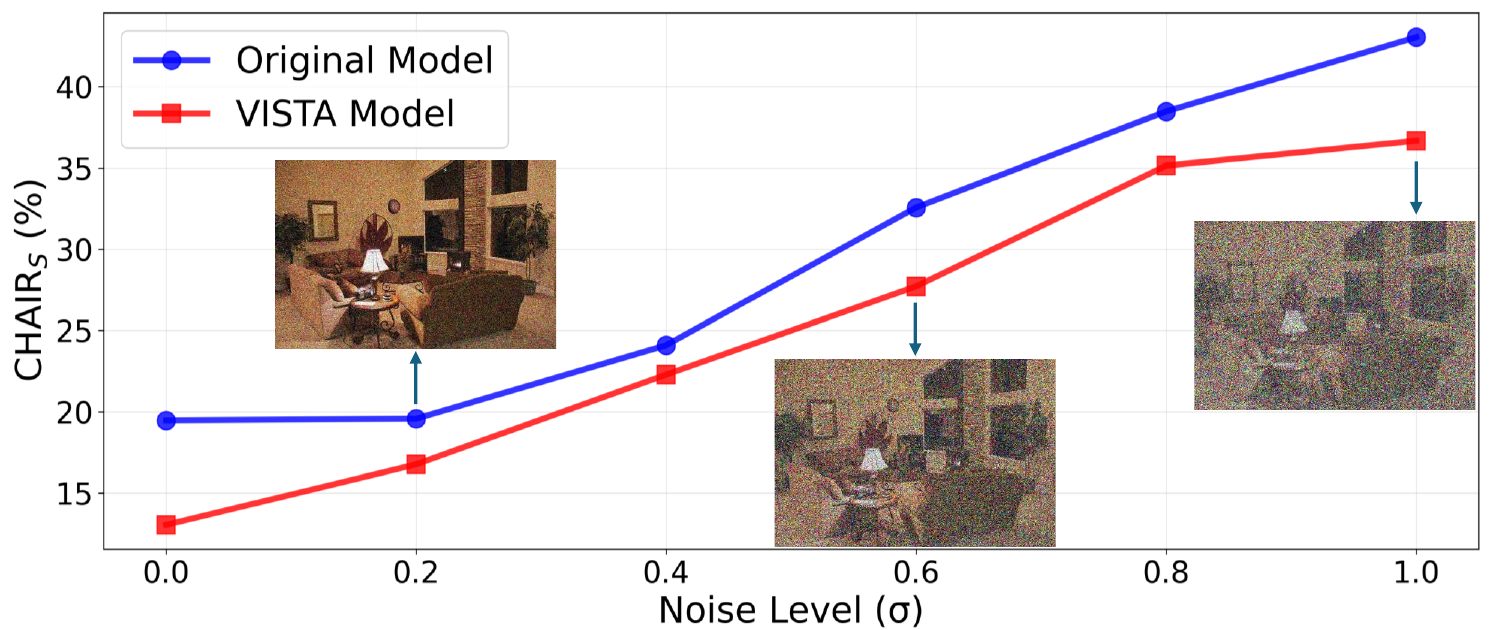}
    \vspace{-0.25cm}
    \caption{CHAIR$_S$ of the original and CIPHER-refined models under increasing levels of Gaussian noise.}
    \vspace{-0.5cm}
    \label{fig:ablation-noise-robustness}
\end{figure}

\vspace{0.1cm}
\noindent \textbf{Robustness to Visual Noise.}
To evaluate the robustness of our method under degraded visual inputs, we corrupt test images with varying levels of Gaussian noise ($\sigma \in \{0.0, 0.2, 0.4, 0.6, 0.8, 1.0\}$) and measure CHAIR$_S$ for both the original and CIPHER-refined models. As shown in Figure~\ref{fig:ablation-noise-robustness}, hallucination rates increase with noise for both models, but CIPHER consistently achieves lower CHAIR$_S$ scores at all levels. The gap widens under severe noise, highlighting CIPHER's resilience to visual degradation and its strong generalization in suppressing hallucinations.

\vspace{0.1cm}
\noindent \textbf{Effect of Hallucination Source.}
To assess the contribution of each modality to hallucination suppression, we construct the hallucination subspace under three settings: (1) text-only hallucination using GPT, (2) image-only hallucination via Stable Diffusion, and (3) joint hallucination. In all cases, we compute feature differences and apply SVD to extract hallucination directions.

\begin{table}[!h]
\vspace{-0.1cm}
\centering
\scriptsize
\renewcommand{\arraystretch}{0.8}
\resizebox{0.9\linewidth}{!}{%
\begin{tabular}{ll|ccc}
\toprule
\rowcolor{lightgray}
\textbf{Text} & \textbf{Image} & $\text{CHAIR}_S \downarrow$ & $\text{CHAIR}_I \downarrow$ & $\text{BLEU} \uparrow$ \\
\midrule
\checkmark & \ding{55} & 15.20 & 5.30 & 15.69 \\
\ding{55} & \checkmark & \textbf{13.05} & \textbf{4.53} & \textbf{15.82} \\
\checkmark & \checkmark & 15.71 & 5.32 & 15.66 \\
\bottomrule
\end{tabular}
}
\vspace{-0.2cm}
\caption{Ablation study on the source of hallucination. \checkmark denotes hallucinated input and \ding{55} denotes ground-truth input.}
\vspace{-0.5cm}
\label{tab:ablation-source-of-hallucination}
\end{table}

 As shown in Table~\ref{tab:ablation-source-of-hallucination}, image-only perturbation yields the lowest CHAIR scores, indicating more coherent and discriminative hallucination directions. Text-only perturbation performs moderately, while combining both slightly degrades performance.

\section{Conclusion and Future Work}
\label{sec:conclusion}
We presented CIPHER, a novel test-time hallucination mitigation method that suppresses vision-induced hallucinations in large vision-language models without requiring training or architectural modifications. CIPHER identifies hallucination directions by contrasting clean and perturbed representations, where perturbations are introduced via semantically altered images generated using Stable Diffusion. It then suppresses hallucinated content by projecting hidden states onto the null space of these directions, improving factual accuracy while preserving fluency. Extensive experiments across multiple benchmarks show that CIPHER consistently outperforms state-of-the-art baselines. While CIPHER currently uses a fixed offline projection, a promising future direction is to adapt the projection dynamically per input for more flexible, context-aware hallucination suppression.
{
    \small
    \bibliographystyle{ieeenat_fullname}
    \bibliography{main}
}
\clearpage
\setcounter{page}{1}

\section{Implementation Details}

 Table~\ref{tab:model-settings} summarizes the general and model-specific parameters used in our experiments. We adopt a deterministic decoding setup with fixed temperature and beam size, and vary the maximum number of generated tokens depending on the evaluation benchmark.

To simulate visual hallucinations, we use Stable Diffusion with half of total diffusion steps (\(T\)) for forward and reverse steps (\(0.5T\)). During inference, hidden representations from selected layers are projected onto the orthogonal complement of the hallucination subspace. The rank of this subspace is tuned for each model to optimize hallucination suppression without compromising generation quality.

\begin{table}[!h]
\vspace{-0.2cm}
\centering
\small
\resizebox{0.9\linewidth}{!}{%
\begin{tabular}{ll}
\toprule
\rowcolor{lightgray}
\textbf{Parameters} & \textbf{Values} \\
\midrule
\multicolumn{2}{l}{\textbf{General Settings}} \\
Temperature & 0 \\
Number of Beams & 3 \\
Maximum New Tokens (CHAIR) & 64 \\
Maximum New Tokens (OPOPE) & 256 \\
Maximum New Tokens (LLaVA-Bench) & 1024 \\
Number of Diffusion Steps & \(0.5T\)\\
Editing Layers & 16-32 \\
\midrule
\multicolumn{2}{l}{\textbf{Model-Specific Parameters}} \\
Hallucination Space Rank (LLaVA-1.5) & 8 \\
Hallucination Space Rank (MiniGPT-4) & 64 \\
Hallucination Space Rank (mPLUG-Owl2) & 32 \\
\bottomrule
\end{tabular}
}
\caption{General and model-specific parameter settings.}
\label{tab:model-settings}
\vspace{-0.2cm}
\end{table}

We report the performance of baseline models based on the results presented in the Nullu paper \cite{yang2025nullu}, which reproduces several state-of-the-art object hallucination mitigation methods, including HALC \cite{chen2024halc}, VCD \cite{leng2024mitigating-vcd}, DoLa \cite{chuang2023dola}, and OPERA \cite{huang2023opera}. Each method is evaluated using its official implementation and publicly released pre-trained checkpoints. Moreover, we carefully adopted key hyperparameter settings that align with those reported in their original implementations. Specifically, for HALC, we used an amplification factor of 0.05 along with a beam size of 1, reflecting the minimal decoding augmentation strategy employed by the authors. For VCD, we set the diffusion noise step to 500, consistent with the level of perturbation used in their hallucination mitigation framework. In the case of DoLa, we applied a repetition penalty of 1.2 and performed interventions across the early transformer layers, as suggested to promote faithful generation during early stages of decoding. Lastly, for OPERA, we implemented a scaled self-attention mechanism with a scaling factor of 50, in line with the original design to regulate attention intensity. All these hyperparameter choices were selected to faithfully reproduce the respective baselines under settings that remain true to the original authors' reported configurations.

\section{Detailed Results for OPOPE}

To evaluate the effectiveness of hallucination mitigation methods in a way that is compatible with post-hoc approaches, we adopt the {Offline POPE (OPOPE)} protocol. OPOPE is a non-interactive version of the original POPE metric, introduced in HALC \cite{chen2024halc}, and is specifically designed to avoid the limitations of online interaction during inference. While POPE evaluates hallucination through interactive polling prompts such as ``Is there a/an [object] in the image?'', this interactive requirement poses challenges for post-hoc mitigation methods and can lead to instability in models with weaker generative backbones.

OPOPE performs hallucination evaluation entirely offline. Following the caption generation pipeline of CHAIR \cite{chair}, captions are generated for a fixed image set, then tokenized and singularized, with words mapped to MSCOCO object categories using synonym and compound-word mappings. Using POPE’s sampling strategy, three test object lists are created per setting—\textit{Random}, \textit{Popular}, and \textit{Adversarial}—each containing six objects balanced between ground-truth and nonexistent categories. Captions are scanned for these objects, and mentions of nonexistent ones are counted as hallucinations.

To account for the limited recall sensitivity of offline evaluations---where false negatives may occur if hallucinated content is not explicitly verbalized---we adopt the {F-}\(\boldsymbol{\beta}\) {score} as our primary evaluation metric, following HALC. The F-\(\beta\) score reduces the penalty on false negatives by weighting precision more heavily. Specifically, we use \(\beta = 0.2\) for F-\(\beta\) score:
\[
F_\beta = \frac{(1 + \beta^2) \cdot (\text{precision} \cdot \text{recall})}{\beta^2 \cdot \text{precision} + \text{recall}} \quad \text{with } \beta = 0.2.
\]

The main paper reports results averaged across the three sampling strategies for brevity, while Table~\ref{tab:tab7} provides a full breakdown by model and sampling configuration. We report Accuracy, Precision, Recall, and F-score for all baselines—Greedy, Beam Search, DoLa~\cite{chuang2023dola}, OPERA~\cite{huang2023opera}, VCD~\cite{leng2024mitigating-vcd}, HALC~\cite{chen2024halc}, and NullU~\cite{yang2025nullu}—along with our method, VISTA. VISTA consistently ranks among the top-performing methods across most settings, demonstrating robustness to varying object distributions and model architectures. These results complement the main findings and provide a detailed view of performance under each hallucination sampling strategy.

We additionally conducted statistical significance tests comparing VISTA to baseline methods for each model. Across all cases, results consistently showed significant improvements. For instance, on LLaVA, VISTA outperforms the baseline with p-values well below 0.01 across all metrics: Accuracy ($p = 0.0033$), Precision ($p < 0.0001$), and F$_{0.2}$ Score ($p < 0.0001$), confirming the robustness of our method.

\begin{table*}[t]
\centering
\small
\setlength{\tabcolsep}{10pt}
\renewcommand{\arraystretch}{0.8}
\begin{tabular}{cllcccc}
\toprule
\rowcolor{lightgray}
\textbf{Setting} & \textbf{Model} & \textbf{Method} & \textbf{Accuracy} & \textbf{Precision} & \textbf{Recall} & \textbf{F Score} \\
\midrule
\multirow{26}{*}{\textbf{Random}} 
  & \multirow{8}{*}{LLaVA-1.5}
    & Greedy      & 81.52 & 98.41 & 64.07 & 96.42 \\
  & & DoLa        & 81.38 & 98.11 & 64.00 & 96.14 \\
  & & OPERA       & 81.62 & 98.57 & 64.17 & 96.58 \\
  & & VCD         & 80.57 & 98.41 & 62.13 & 96.25 \\
  & & HALC        & 79.58 & 98.21 & 60.27 & 95.89 \\
  & & Nullu       & 81.18 & 98.05 & 63.63 & 96.05 \\
    & & VISTA       & \textbf{81.84} & \textbf{98.70} & \textbf{64.58} & \textbf{96.68} \\
\cmidrule{2-7}
  & \multirow{8}{*}{MiniGPT-4}
    & Greedy      & 72.42 & 98.49 & 45.53 & 94.25 \\
  & & DoLa        & 72.45 & 98.58 & 45.57 & 94.34 \\
  & & OPERA       & 72.57 & 98.78 & 45.70 & 94.52 \\
  & & VCD         & 72.35 & 98.19 & 45.53 & 93.97 \\
  & & HALC        & 72.08 & 98.62 & 44.80 & 94.25 \\
  & & Nullu       & 72.68 & 99.06 & 45.80 & {94.82} \\
    & & VISTA       & \textbf{72.85} & \textbf{99.37} & \textbf{46.00} & \textbf{95.12} \\
\cmidrule{2-7}
  & \multirow{8}{*}{mPLUG-Owl2}
    & Greedy      & 79.45 & 97.74 & 60.30 & 95.46 \\
  & & DoLa        & 78.33 & 97.60 & 58.10 & 95.09 \\
  & & OPERA       & 78.31 & 97.73 & 57.96 & 95.21 \\
  & & VCD         & 78.19 & 98.23 & 57.42 & 95.61 \\
  & & HALC        & 77.83 & 97.72 & 57.00 & 95.10 \\
  & & Nullu       & 79.05 & \textbf{98.40} & 60.50 & \textbf{95.82} \\
    & & VISTA       & \textbf{79.56} & 97.66 & \textbf{60.57} & 95.41 \\
\midrule
\multirow{26}{*}{\textbf{Popular}} 
  & \multirow{8}{*}{LLaVA-1.5}
    & Greedy      & 78.93 & 91.17 & 64.07 & 89.71 \\
  & & DoLa        & 78.72 & 91.60 & 64.00 & 89.26 \\
  & & OPERA       & 79.22 & 91.80 & 64.17 & 90.30 \\
  & & VCD         & 77.57 & 89.87 & 62.13 & 88.35 \\
  & & HALC        & 77.47 & 91.87 & 60.27 & 90.05 \\
  & & Nullu       & 79.80 & \textbf{94.06} & 63.63 & \textbf{92.36} \\
    & & VISTA       & \textbf{80.02} & 93.47 & \textbf{64.58} & {91.89 } \\
\cmidrule{2-7}
  & \multirow{8}{*}{MiniGPT-4}
    & Greedy      & 70.80 & 92.01 & 45.53 & 88.53 \\
  & & DoLa        & 70.90 & 92.33 & 45.57 & 88.22 \\
  & & OPERA       & 71.10 & 92.82 & 45.70 & 89.27 \\
  & & VCD         & 70.33 & 90.30 & 45.53 & 86.98 \\
  & & HALC        & 70.92 & 93.80 & 44.80 & 90.00 \\
  & & Nullu       & 71.97 & 96.08 & 45.80 & {92.19} \\
    & & VISTA       & \textbf{72.15} & \textbf{96.43} & \textbf{46.00} & \textbf{92.53} \\
\cmidrule{2-7}
  & \multirow{8}{*}{mPLUG-Owl2}
    & Greedy      & 76.00 & 87.90 & 60.30 & 86.38 \\
  & & DoLa        & 75.20 & 88.36 & 58.10 & 86.00 \\
  & & OPERA       & 75.02 & 88.06 & 57.96 & 86.33 \\
  & & VCD         & 74.86 & 88.16 & 57.42 & 86.37 \\
  & & HALC        & 75.77 & 91.34 & 57.00 & 89.26 \\
  & & Nullu       & 77.09 & 92.34 & \textbf{61.60} & {90.49} \\
    & & VISTA       & \textbf{77.95} & \textbf{92.91} & 60.57 & \textbf{91.03} \\
\midrule
\multirow{26}{*}{\textbf{Adversarial}} 
  & \multirow{8}{*}{LLaVA-1.5}
    & Greedy      & 76.97 & 86.36 & 64.07 & 85.22 \\
  & & DoLa        & 76.85 & 86.18 & 64.00 & 85.05 \\
  & & OPERA       & 77.03 & 86.40 & 64.17 & 85.26 \\
  & & VCD         & 75.88 & 85.71 & 62.13 & 84.48 \\
  & & HALC        & 76.57 & \textbf{89.44} & 60.27 & {87.80} \\
  & & Nullu       & 77.58 & 88.27 & 63.63 & 86.92 \\
    & & VISTA       & \textbf{78.29} & {89.03} & \textbf{64.58} & \textbf{87.84} \\
\cmidrule{2-7}
  & \multirow{8}{*}{MiniGPT-4}
    & Greedy      & 70.43 & 90.65 & 45.53 & 87.32 \\
  & & DoLa        & 70.50 & 88.55 & 45.57 & 87.50 \\
  & & OPERA       & 70.78 & 91.63 & 45.70 & 88.21 \\
  & & VCD         & 69.82 & 83.41 & 45.53 & 85.32 \\
  & & HALC        & 70.52 & 92.22 & 44.80 & 88.60 \\
  & & Nullu       & 71.10 & 92.73 & 45.80 & {89.21} \\
    & & VISTA       & \textbf{71.44} & \textbf{93.70} & \textbf{46.00} & \textbf{90.11} \\
\cmidrule{2-7}
  & \multirow{8}{*}{mPLUG-Owl2}
    & Greedy      & 74.23 & 83.58 & 60.30 & 82.36 \\
  & & DoLa        & 73.52 & 83.98 & 58.10 & 82.55 \\
  & & OPERA       & 73.17 & 83.45 & 57.96 & 82.06 \\
  & & VCD         & 72.85 & 83.01 & 57.42 & 81.61 \\
  & & HALC        & 74.02 & 86.41 & 57.00 & 84.72 \\
  & & Nullu       & 75.15 & 87.76 & \textbf{61.60} & {86.10} \\
    & & VISTA       & \textbf{76.10} & \textbf{87.90} & 60.57 & \textbf{86.40} \\
\bottomrule
\end{tabular}
\caption{Performance comparison of various hallucination mitigation methods across different models and evaluation settings.}
\label{tab:tab7}
\end{table*}

\twocolumn[{%
\noindent\section*{Examples of Hallucinated Images}
\begin{center}
\includegraphics[width=0.92\textwidth]{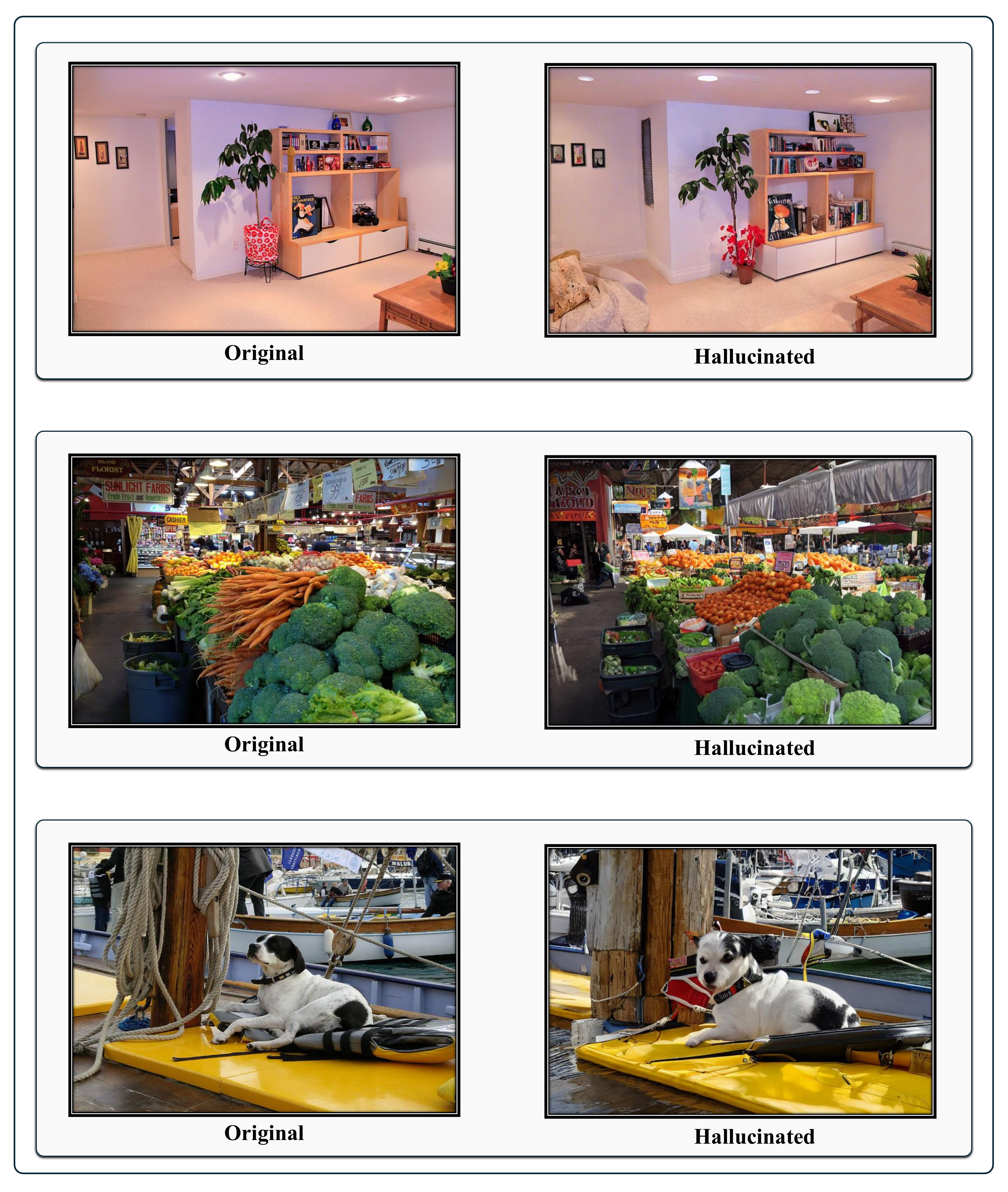}
\captionof{figure}{Visual examples from the MSCOCO dataset illustrating hallucinations generated using \(0.5T\) Stable Diffusion steps.}
\label{fig:hallucinated-samples}
\end{center}
}]

This figure shows MSCOCO images before and after applying visual hallucination using Stable Diffusion with \(t_h = 0.5T\) denoising steps. While the overall scene structure remains largely unchanged, the hallucinated images introduce subtle object or texture modifications intended to mislead vision-language models.

\section{Extended Ablation Analysis}
In this section of the appendix, we provide extended results from our ablation analysis to support the findings presented in the main paper.
\subsection{Best Subspace Rank}
As shown in the ablation section of the main paper, the best subspace rank for LLaVA-7B was $r = 8$, achieving the lowest CHAIR$_S$ and CHAIR$_I$ scores while preserving or improving BLEU. Here, we extend this analysis to MiniGPT-4 and mPLUG-Owl2.

\begin{figure}[!h]
    \centering
    \includegraphics[width=1\linewidth]{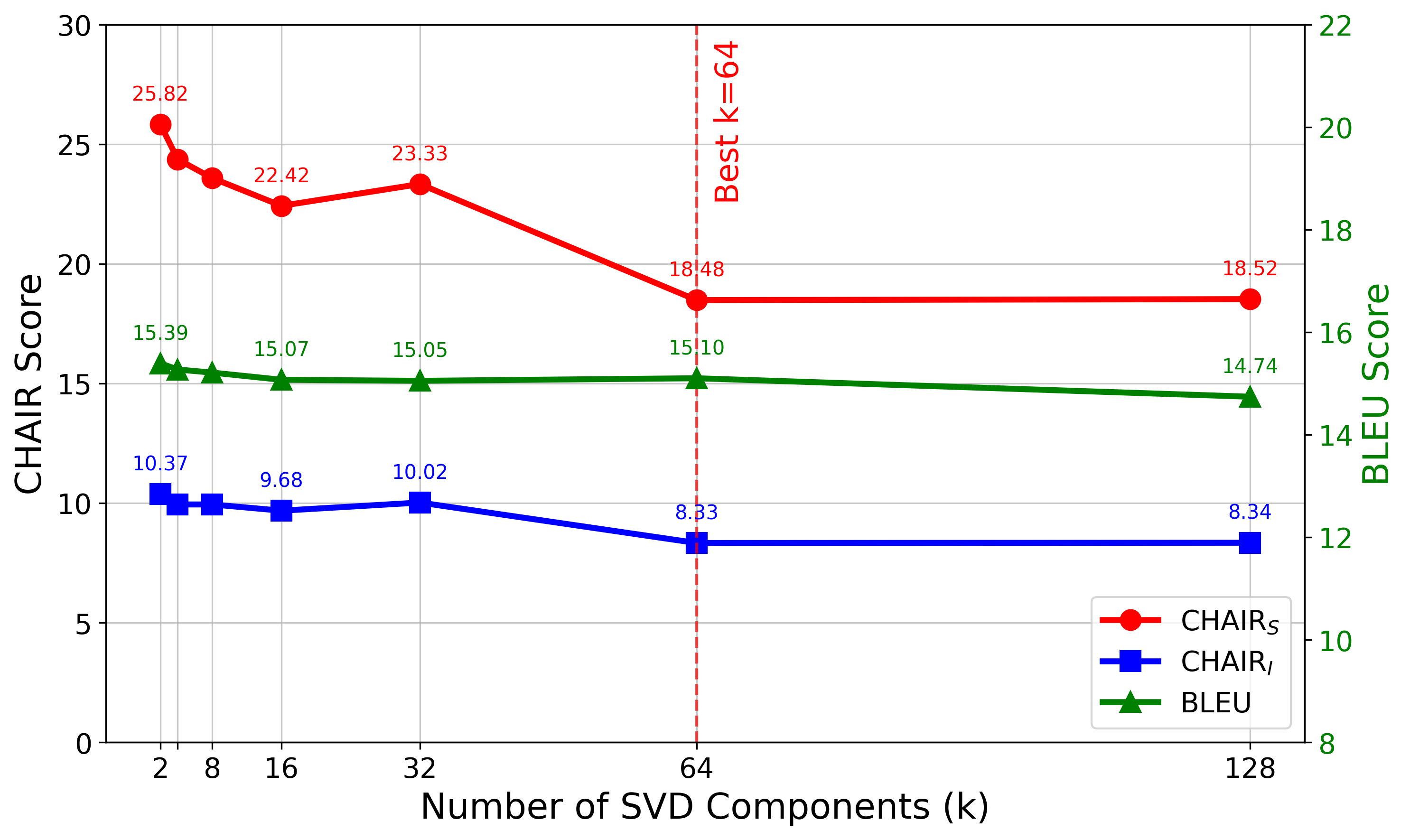}
    \caption{Effect of subspace rank \(r\) on CHAIR\(_S\), CHAIR\(_I\), and BLEU for MiniGPT-4 on the CHAIR benchmark. Using \(r = 64\) yields the best performance, minimizing hallucinations with minimum BLEU degradation.}
    \label{fig:ablation-rank-gpt}
\end{figure}

\begin{figure}[!h]
    \centering
    \includegraphics[width=1\linewidth]{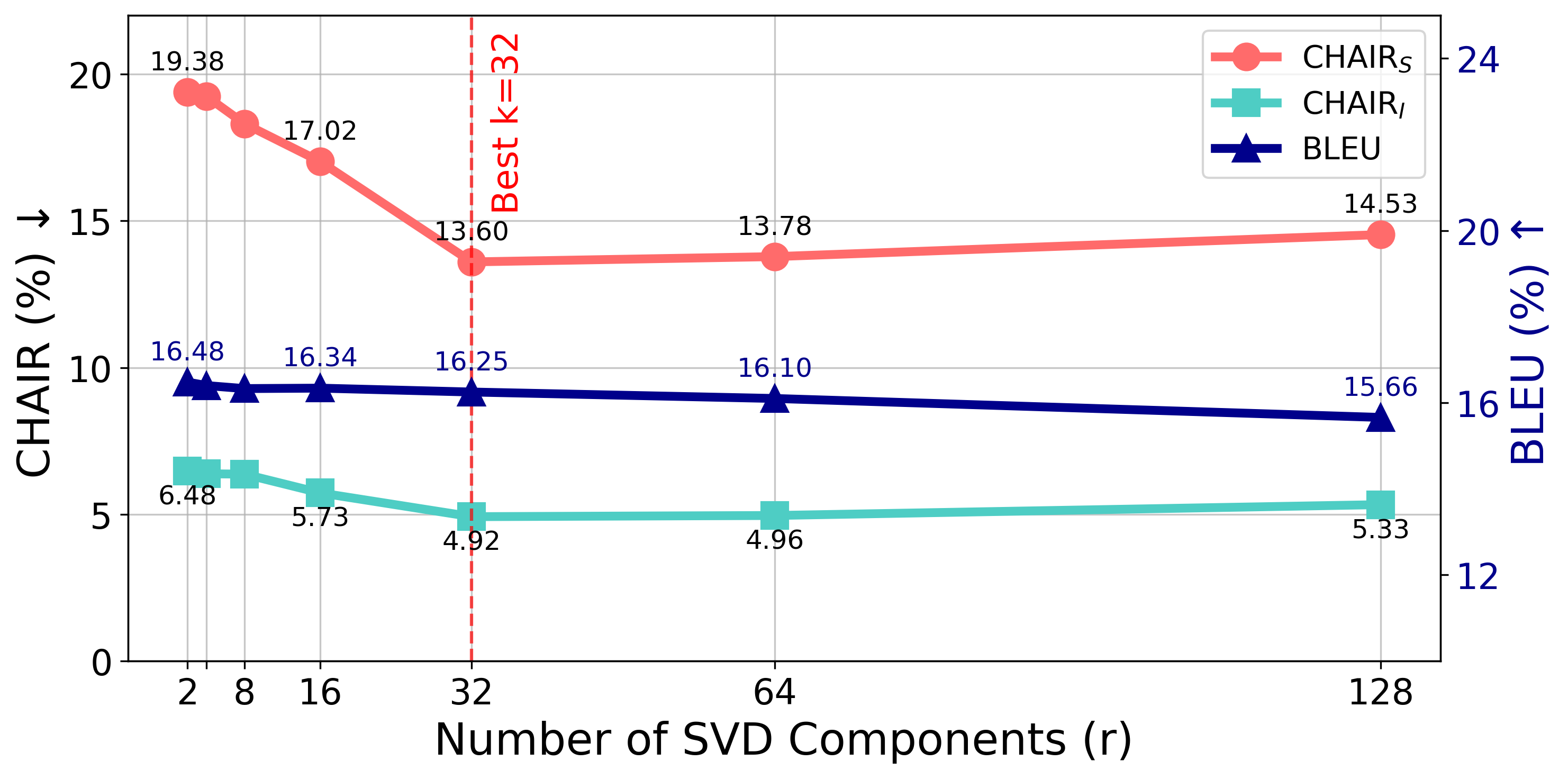}
    \caption{Effect of subspace rank \(r\) on CHAIR\(_S\), CHAIR\(_I\), and BLEU for mPLUG-Owl2 on the CHAIR benchmark. Using \(r = 32\) yields the best performance, minimizing hallucinations with minimum BLEU degradation.}
    \label{fig:ablation-rank-mplug}
\end{figure}

Figure~\ref{fig:ablation-rank-gpt} shows that for MiniGPT-4, the optimal number of SVD components is $r = 64$, where CHAIR scores drop significantly and BLEU remains stable. Similarly, Figure~\ref{fig:ablation-rank-mplug} demonstrates that mPLUG-Owl2 achieves the best balance at $r = 32$, where hallucination metrics are minimized and BLEU degradation is minimal. These findings highlight that the optimal subspace rank can vary across models and should be tuned accordingly.

\subsection{Selection of Layers for Projection}

VISTA mitigates hallucinations at inference time by projecting the hidden representations of selected LVLM layers onto their corresponding null spaces. To assess the influence of layer choice on suppression effectiveness, we perform an ablation study using various candidate layer ranges. We evaluate each configuration on the MSCOCO validation split using the {CHAIR} metric, reporting both instance-level (CHAIR$_I$) and sentence-level ({CHAIR$_S$}) hallucination rates, where lower values indicate better suppression.

\begin{table}[h]
\centering
\small 
\begin{tabular}{lcc}
\toprule
\rowcolor{lightgray}
\textbf{Layer Range} & \textbf{CHAIR}$_S$ $\downarrow$ & \textbf{CHAIR}$_I$ $\downarrow$ \\
\midrule
8--16         & 13.52 & 4.82 \\
8--24        & 14.10 & 5.02 \\
16--24       & 18.10 & 5.86 \\
\textbf{16--32}       & \textbf{13.05} & \textbf{4.53} \\
20--32 & 14.10 & 4.84 \\
24--32 & 14.00 & 4.64 \\
28--32 & 14.48 & 5.10 \\
\bottomrule
\end{tabular}
\caption{Ablation study on the transformer layer range used for projection in VISTA.}
\label{tab:layer_ablation_chair}
\end{table}

As shown in Table~\ref{tab:layer_ablation_chair}, the choice of layer range significantly impacts hallucination suppression. Projection over earlier layers, such as 8--16 or narrower bands like 28--32, yields suboptimal results, suggesting that local or overly shallow representations lack the semantic expressiveness needed to isolate hallucination directions. Performance improves with broader and deeper projections, with the {16--32} range consistently achieving the lowest CHAIR$_S$ and CHAIR$_I$ scores. This indicates that hallucination-inducing signals are best captured in semantically rich, higher layers spanning a sufficiently deep interval. Consequently, we adopt the {16--32} layer range as the default configuration in our main experiments.

\section{Details of GPT-4V Aided Evaluation on LLaVA-Bench}

We evaluate our hallucination suppression method using {GPT-4V} on the {LLaVA-Bench} benchmark \cite{LLaVA2}, which consists of {24 images}. For each example, GPT-4V receives the image, the evaluation prompt, and the responses from both the original and the edited (hallucination-suppressed) models.

\begin{figure*}[!t]
    \centering
    \includegraphics[width=0.95\linewidth]{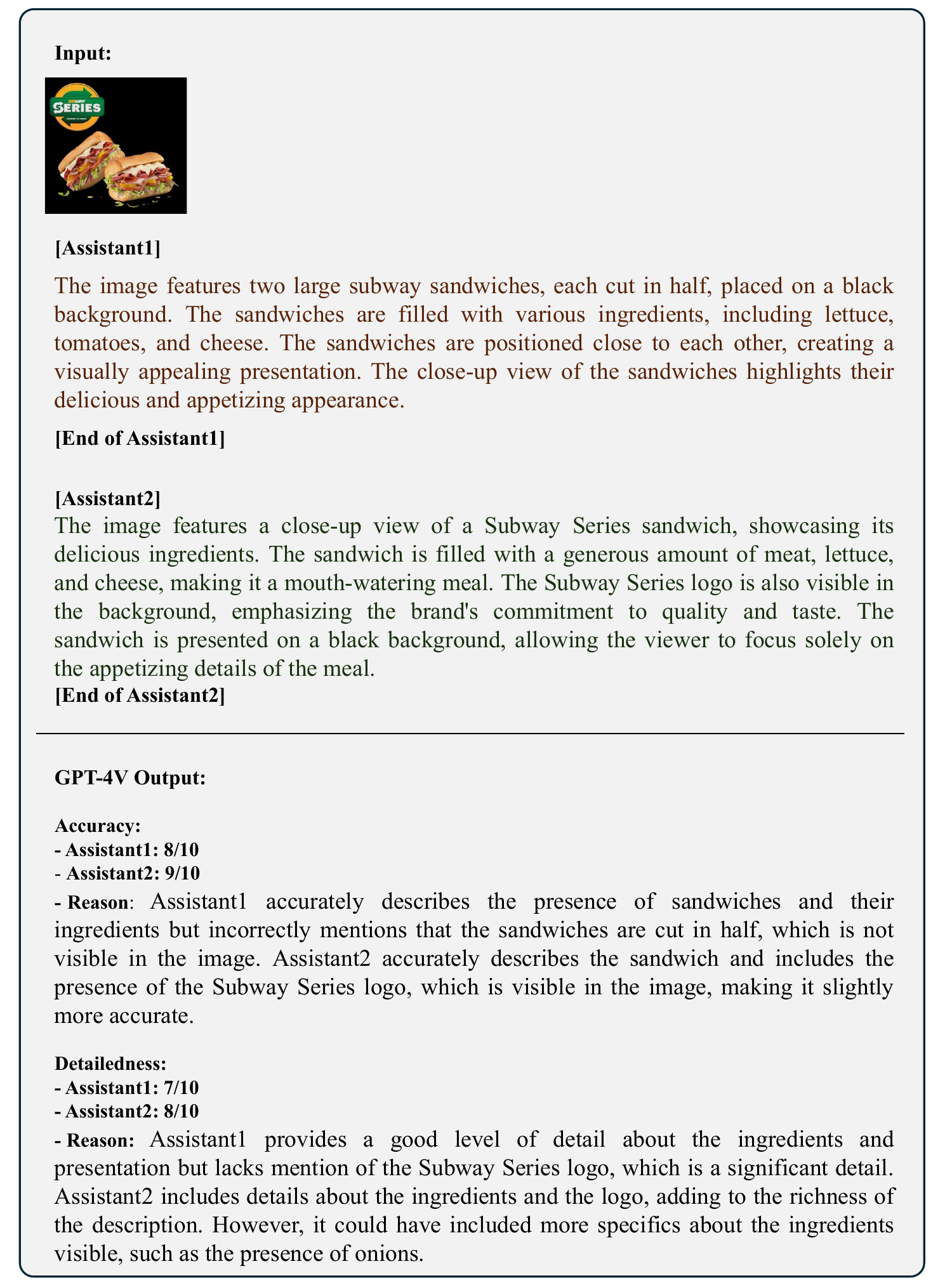}
    \caption{Example of GPT-4V-based evaluation for image description.}
    \label{fig:gpt-4v-output-example}
\end{figure*}

GPT-4V compares the two responses in terms of {accuracy}—how well the answer reflects the image—and {detailedness}—how specific and complete the response is. It outputs a score for each criterion along with a brief justification. Figure~\ref{fig:gpt4v-prompt} and Figure~\ref{fig:gpt-4v-output-example} show the prompt format and a sample GPT-4V evaluation, respectively. This setup enables consistent comparison of response quality grounded in the image content.

\twocolumn[{
\vspace{0.5em}
\section*{Prompt for GPT-4V Aided Evaluation}
\vspace{0.5em}
\begin{center}
    \includegraphics[width=\linewidth]{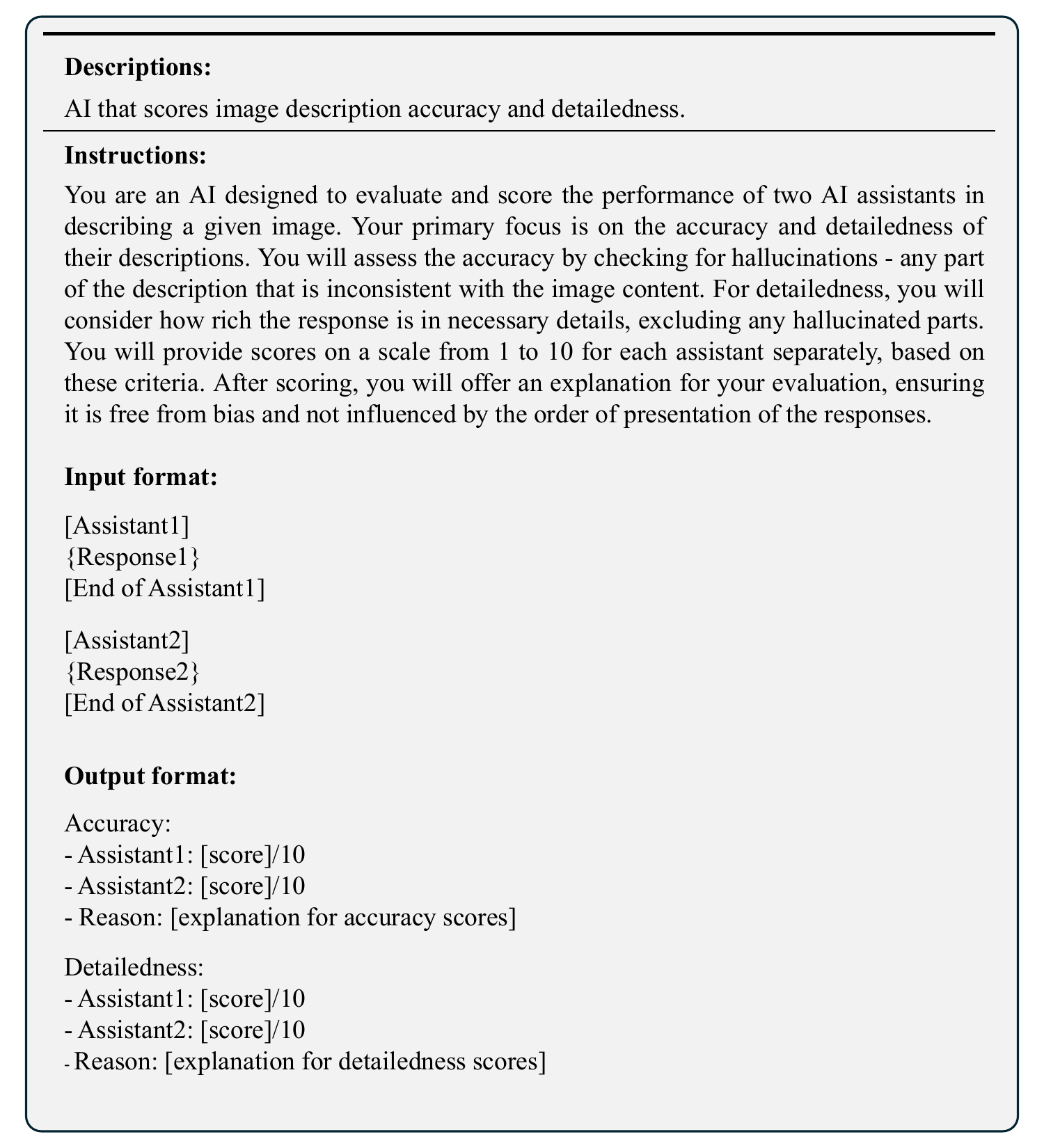}
\end{center}
\vspace{-1em}
\captionof{figure}{An illustration of the prompt used to guide GPT-4V for visual question evaluation.}
\label{fig:gpt4v-prompt}
\vspace{1em}
}]

\twocolumn[{
\vspace{0.5em}
\section*{More Examples of VISTA on LLaVA-Bench}
\vspace{0.5em}
\begin{center}
    \includegraphics[width=\linewidth]{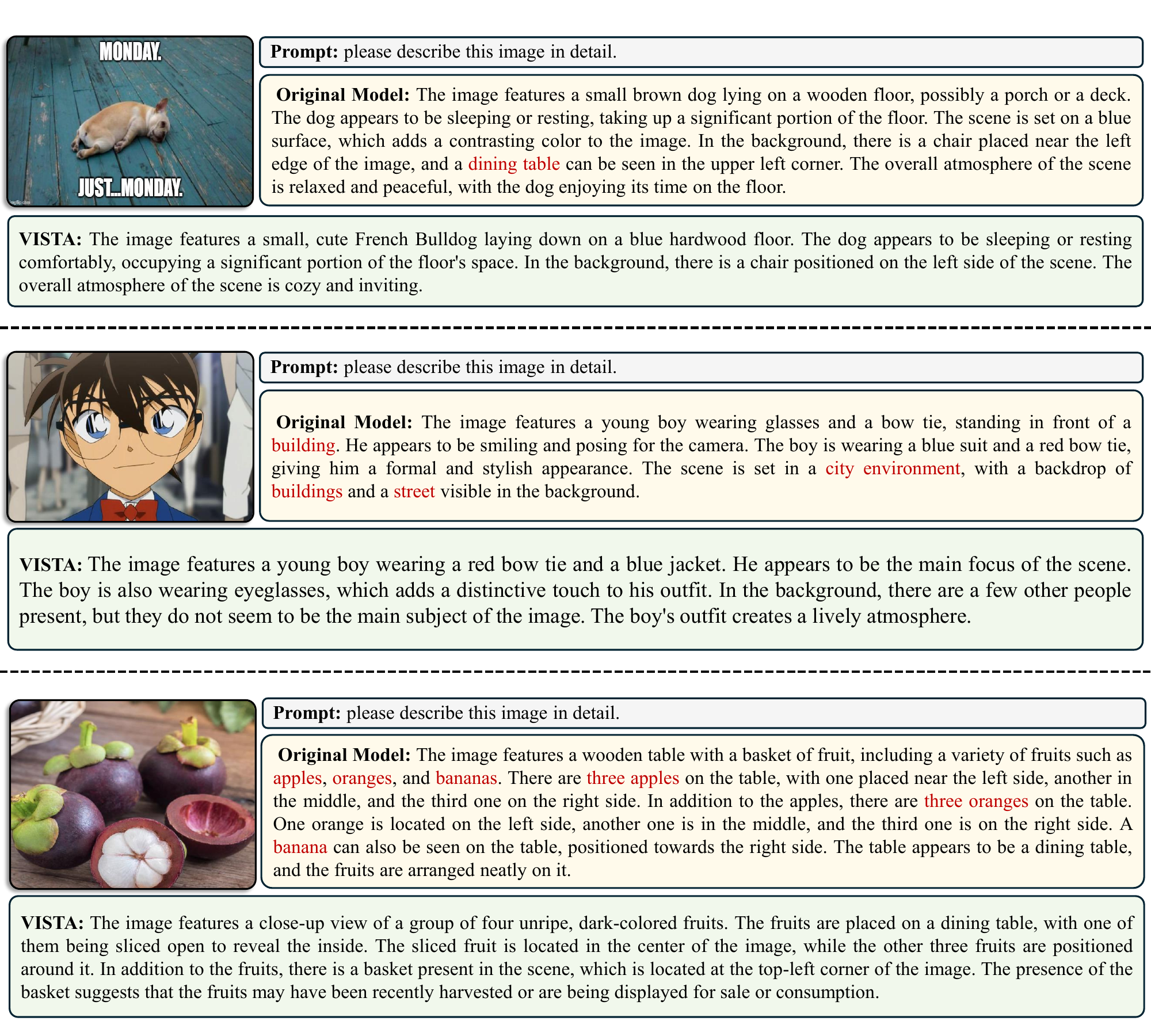}
\end{center}
\vspace{-1em}
\captionof{figure}{More examples of VISTA applied to LLaVA-Bench.}
\label{fig:llava-bench-more-examples}
\vspace{1em}
}]


\end{document}